\documentclass[runningheads]{llncs}

\usepackage[final,year=2024,ID=12638]{eccv}

\usepackage{eccvabbrv}
\usepackage{wrapfig}

\usepackage{multirow}

\usepackage{graphicx}
\usepackage{booktabs}

\usepackage[accsupp]{axessibility}  

\usepackage{hyperref}

\usepackage{orcidlink}

\newcommand{\fullModelWarmup}{{FM-Warmup}\xspace}
\newcommand{\fullModelWarmupSym}{{$P^{fm}_{\text{warmup}}$}\xspace}

\newcommand{\LRadjust}{{$\mathcal{E}$-Shrinking}\xspace}

\newcommand{\proposedinplaceKD}{{IKD-Warmup}\xspace}
\newcommand{\proposedTraining}{D$\epsilon$pS\xspace}

\newcommand{\ofa}{OFA\xspace}
\newcommand{\epsshrinking}{$\mathcal{E}$-Shrinking\xspace}
\newcommand{\ofaCite}{\ofa\cite{ofa}\xspace}
\newcommand{\bignas}{BigNAS\xspace}
\newcommand{\bignasCite}{\bignas\cite{bignas}\xspace}

\newcommand{\tableref}[1]{Tab.~\ref{#1}}
\newcommand{\secref}[1]{\S\ref{#1}} 
\newcommand{\figref}[1]{Fig. \ref{#1}} 

\newenvironment{tightitemize}{%
\begin{list}{$\bullet$}{%
\setlength{\itemsep}{1.5pt}%
\setlength{\topsep}{2pt}%
\setlength{\parskip}{0pt}%
\setlength{\parsep}{0pt}%

\setlength{\labelwidth}{0pt}%
\setlength{\leftmargin}{4pt}%
\setlength{\labelsep}{0pt}%
\setlength{\listparindent}{0pt}%
}}%
{\end{list}}

\begin{document}

\definecolor{codegreen}{rgb}{0,0.6,0}
\definecolor{purple}{RGB}{128,0,128}
\definecolor{indigo}{RGB}{75,0,130}
\definecolor{royalblue}{RGB}{65,105,225}
\definecolor{navy}{RGB}{0,0,128}
\definecolor{codebrown}{rgb}{0.6,0.6,0}

\newif\ifcommenton
\commentontrue
\ifcommenton

\newcommand{\alind}[1]{\textcolor{codegreen}{[Alind: #1]}}
\newcommand{\alexey}[1]{\textcolor{purple}{[AT: #1]}}
\newcommand{\aditya}[1]{\textcolor{indigo}{[Aditya: #1]}}
\newcommand{\hugo}[1]{\textcolor{royalblue}{[Hugo: #1]}}
\newcommand{\myungjin}[1]{\textcolor{codebrown}{[Myungjin: #1]}}
\newcommand{\igor}[1]{\textcolor{navy}{[Igor: #1]}}

\else
\newcommand{\aditya}[1]{}
\newcommand{\alind}[1]{}
\newcommand{\alexey}[1]{}
\newcommand{\hugo}[1]{}
\newcommand{\myungjin}[1]{}
\newcommand{\igor}[1]{}

\fi

\title{D$\epsilon$pS: Delayed $\epsilon$-Shrinking for Faster Once-For-All Training} 

\titlerunning{D$\epsilon$pS: Delayed $\epsilon$-Shrinking}

\author{Aditya Annavajjala$^{*}$\inst{1} \and Alind Khare$^{*}$\inst{1} \and Animesh Agrawal\inst{1} \and Igor Fedorov\inst{3} \and Hugo Latapie\inst{2} \and Myungjin Lee\inst{2} \and Alexey Tumanov\inst{1}}


\authorrunning{Annavajjala et al.}

\institute{Georgia Institute of Technology, Atlanta, USA \and
Cisco Research, USA \and Meta, USA}


\maketitle
\def\thefootnote{$*$}\footnotetext{Authors contributed equally to this research.}
\begin{abstract}

CNNs are increasingly deployed across different hardware, dynamic environments, and low-power embedded devices. This has led to the design and training of CNN architectures with the goal of maximizing accuracy subject to such variable deployment constraints. As the number of deployment scenarios grows, there is a need to find scalable solutions to design and train specialized CNNs.
Once-for-all training  has emerged as a scalable approach that jointly co-trains many models (subnets) at once with a constant training cost and finds specialized CNNs later. The scalability is achieved by training the full model and simultaneously reducing it to smaller subnets that share model weights (weight-shared shrinking). However, existing once-for-all training approaches incur huge training costs reaching 1200 GPU hours. 
We argue this is because they either start the process of shrinking the full model \textit{too} early or \textit{too} late. Hence, we propose Delayed $\mathcal{E}$-Shrinking (\proposedTraining) that starts the process of shrinking the full model when it is \textit{partially} trained ($\sim 50\%$) which leads to training cost improvement and better in-place knowledge distillation to smaller models.
The proposed approach also consists of novel heuristics that dynamically adjust subnet learning rates incrementally ($\mathcal{E}$), leading to improved weight-shared knowledge distillation from larger to smaller subnets as well.
As a result, \proposedTraining outperforms state-of-the-art once-for-all training techniques across different datasets including CIFAR10/100, ImageNet-100, and ImageNet-1k on accuracy and cost.
It achieves $1.83\%$ higher ImageNet-1k top1 accuracy or the same accuracy with $1.3$x reduction in FLOPs and   $2.5$x drop in training cost (GPU*hrs). 

\end{abstract}

\section{Introduction}
\label{sec:intro}

CNNs are pervasive in numerous applications including smart cameras \cite{smart_camera}, smart surveillance \cite{smart_survelliance}, self-driving cars \cite{self_driving}, search engines \cite{cnn_search_engine}, and social media \cite{tcn}. As a result, they are increasingly deployed across diverse hardware ranging from server-grade GPUs like V100 \cite{nvidia_v100} to edge-GPUs like Nvidia Jetson \cite{nvidia_jetson} and dynamic environments like Autonomous Vehicles \cite{pylot} that operate under strict latency or power budget constraints. As the diversity in deployment scenarios grows, efficient deployment of CNNs on a myriad of deployment constraints becomes challenging. It calls for developing techniques that find appropriate CNNs suited for different deployment conditions.

Neural Architecture Search (NAS) \cite{proxylessnas, mnasnet} has emerged as a successful technique that 
finds CNN architectures specialized for a deployment target.
It searches for appropriate CNN architecture and trains it with the goal of maximizing accuracy subject to deployment constraints. However, state-of-the-art NAS techniques remain prohibitively expensive, requiring many GPU hours due to the costly operation of the search and training of specialized CNNs. The problem is exacerbated when NAS is employed to satisfy multiple deployment targets, as it must be run repeatedly for \textit{each} deployment target. This makes the cost of NAS 
linear in the number of deployment targets considered ($O(k)$), which is prohibitively expensive and doesn't scale with the growing  number of deployment targets.
Therefore, there is a need to develop scalable NAS solutions able to satisfy multiple deployment targets efficiently. 
\\
One such technique is Once-for-all training \cite{ofa, compofa, bignas}---a step towards making NAS computationally feasible to satisfy multiple deployment targets by decoupling training from search.
It achieves this decoupling by co-training a family of models (weight-shared subnets with varied shapes and sizes) embedded inside a supernet \textit{once}, incurring a constant training cost. After the supernet is trained, NAS can be performed for any specific deployment target by simply extracting a specialized subnet from the supernet without retraining (\textit{once-for-all}).
\begin{wrapfigure}{t}{0.5\columnwidth}
    \vspace{-0.15in}
    \includegraphics[width=0.5\columnwidth]{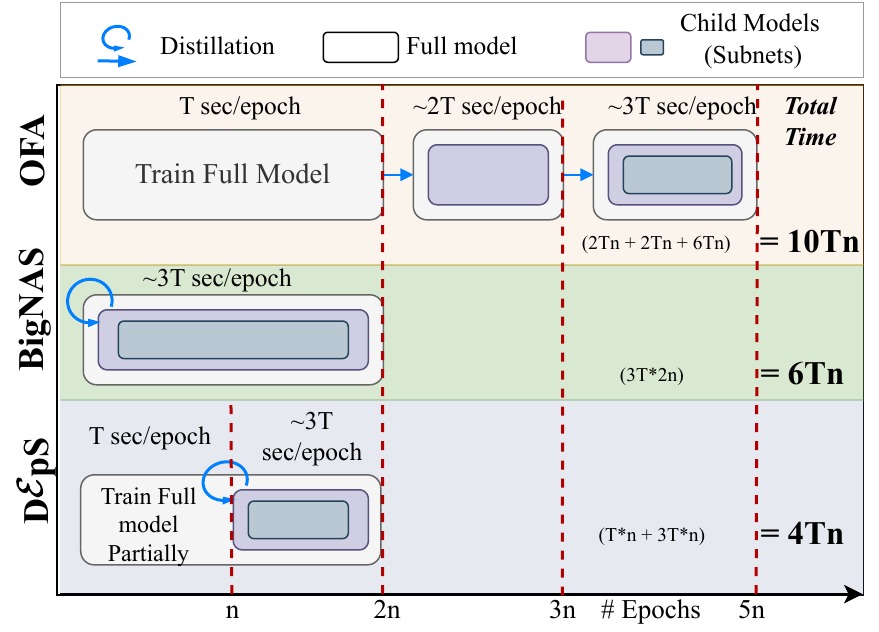}	
	\caption{
	    \small 
        \proposedTraining reduces training time compared to existing approaches like \ofaCite \& \bignasCite.
	}
    \vspace{-0.2in}
    \label{fig:comparison_sota_dss}
\end{wrapfigure}

This achieves $O(1)$ training cost \wrt the number of deployment targets and, therefore, makes NAS scalable. However, the efficiency of this once-for-all training remains limited as it incurs a significant training cost ($\sim$ 1200 GPU hours in \cite{ofa}). This is primarily due to 
(a) the large number of training epochs required to overcome training interference (\ofaCite in \figref{fig:comparison_sota_dss}), and 
(b) the high average time per-epoch caused by shrinking---defined as sampling and adding smaller subnets to the training schedule--- 
per minibatch (\bignasCite in \figref{fig:comparison_sota_dss}).
Thus, in order to make once-for-all training more efficient, we must reduce its training time without sacrificing state-of-the-art accuracy across the whole operating latency/FLOP range of the supernet.

We propose \proposedTraining, a technique that increases the scalability of once-for-all training.
It consists of three key components designed to meet their respective goals --- Full Model warmup (\fullModelWarmup) provides better supernet initialization, \LRadjust keeps the accuracy of the full model (largest subnet that contains all the supernet parameters) on par with OFA and BigNAS, and \proposedinplaceKD boosts the accuracy of small subnets with effective knowledge distillation in once-for-all training. Particularly, with better supernet initialization, \fullModelWarmup (\proposedTraining in \figref{fig:comparison_sota_dss}) reduces both the total number of epochs (compared to OFA) and average time per-epoch (compared to BigNAS). In \fullModelWarmup, the supernet is initialized with the partially trained full model  ($\sim$50\%) and then subnet sampling (shrinking) is started to train the model family. The partial full model training ensures a lower time per epoch initially.
Then, \LRadjust ensures smooth optimization of the full model. It incrementally warms up the learning rate of subnets using parameter $\mathcal{E}$ when the shrinking starts, while keeping the learning rate of the full model higher. Lastly, 
\proposedinplaceKD enables knowledge distillation from multiple partially trained full models (that are progressively better) to smaller subnets. The three components, when combined, reduce the training time of once-for-all training and outperform state-of-the-art \wrt accuracy of subnets across different datasets and neural network architectures. We summarize the contributions of our work as follows:

\begin{tightitemize}
    \item \fullModelWarmup provides better initialization to the weight shared supernet by training the full model only partially and delaying model shrinking. This leads to reduced time per epoch and lower training cost.
    \item \LRadjust ensures smooth and fast optimization of the full model by warming up the learning rate of smaller subnets. This enables it to reach optimal accuracy quickly.
    \item \proposedinplaceKD provides rich knowledge transfer to subnets, enabling them to quickly learn good representations.
\end{tightitemize}

We extensively evaluate \proposedTraining against existing once-for-all training baselines \cite{bignas, ofa, compofa} on CIFAR10/100 \cite{cifar}, ImageNet-100 \cite{imagenet100} as well as ImageNet-1k \cite{imagenet1k} datasets. \proposedTraining outperforms all baselines across all datasets both \wrt accuracy (of subnets) and training cost. It achieves $1.83\%$ ImageNet-1k top1 accuracy improvement or the same accuracy with $1.3$x FLOPs reduction while reducing training cost by upto $1.8$x \wrt OFA and $2.5$x \wrt BigNAS (in dollars or GPU hours). 
We also provide a detailed ablation study to demonstrate the benefits of \proposedTraining components in isolation.

\section{Background}
\label{sec:background}
\textbf{Formulation.} Let $W_o$ denote the supernet's weights, the objective of once-for-all training is given by ---
\begin{equation}\label{eq:problem_formulation}
\underset{W_o}{min} \underset{a \in \mathcal{A}}{\sum} \mathcal{L}(S(W_o, a))
\end{equation}
where $S(W_o, a)$ denotes weights of subnet $a$ selected from the supernet's weight $W_o$ and $\mathcal{A}$ represents the set of all possible neural architectures (subnets). The goal of once-for-all training is to find optimal supernet weights that minimize the loss ($\mathcal{L}$) of all the neural architectures in $\mathcal{A}$ on a given dataset.

\noindent \textbf{Challenges.} However, optimizing \eqref{eq:problem_formulation} is non-trivial. 
On one hand, enumerating gradients of all subnets to optimize the overall objective is computationally infeasible. This is due to the large number of subnets optimized in once-for-all training ($|\mathcal{A}| \approx 10^{19}$ subnets in \cite{ofa}). On the other hand, a naive approximation of objective \eqref{eq:problem_formulation} to make it computationally feasible leads to \textit{interference} (sampling a few subnets in each update step). \textit{Interference} occurs when smaller subnets affect the performance of the larger subnets \cite{ofa, bignas}. Hence, interference causes sub-optimal accuracy of the larger subnets. Existing once-for-all training techniques mitigate interference by increasing the training time significantly (\figref{fig:comparison_sota_dss}). For instance, \ofaCite  mitigates interference by first training the full model (largest subnet) and then progressively increasing the size of $|\mathcal{A}|$. This leads to a large number of training epochs and $\approx 1200$ GPU hours to perform once-for-all training. 
Therefore, the following challenges remain in once-for-all training --- \textbf{(C1)} training supernet at a lesser training cost than SOTA, \textit{and} \textbf{(C2)} mitigating interference. We divide challenge \textbf{C2} into two sub-challenges --- matching existing once-for-all training techniques \cite{ofa, bignas} \wrt accuracy of \textbf{(C2a)}  the full model (largest subnet), and \textbf{(C2b)} child models (smaller subnets).

\section{Related Work}
\label{sec:related_work}
\textbf{Efficient NN-Architectures in Deep Learning.}
Efficient deep neural networks (NNs) achieve high accuracy at low FLOPs. These neural nets are easy to deploy as they increase hardware efficiency by operating at low FLOPs.
Developing such networks is an active research area. Several  efficient neural networks include MobileNets \cite{mbv3}, SqueezeNets \cite{squeezenet}, EfficientNets \cite{efficientnet}, and TinyNets \cite{tinynet}. 

\noindent \textbf{Neural network compression.} Neural network compression reduces the size and computation of neural networks for efficient deployment. The compression occurs after the network is trained. Hence, the performance of compression methods is bounded by the accuracy of the trained neural network. Neural network compression can be broadly divided into two categories --- network pruning and quantization. Network pruning removes  unimportant units \cite{unstructured_pruning_1, unstructured_pruning_2, structured_pruning_3} or channels \cite{structured_pruning_1, structured_pruning_2, unstructured_pruning_3}. Network quantization converts the representation of neural weights and activations to low bits  \cite{quantization_1, quantization_2, quantization_3}. 

\noindent \textbf{Hardware aware NAS.}
Neural architecture search (NAS) automates the design of efficient NN architectures. NAS typically involves searching for and training NN architectures that are more accurate than manually designed NNs  \cite{zoph2018learning, real2019regularized}. Recently, NAS methods are becoming hardware-aware \cite{proxylessnas, mnasnet, fbnet} \ie they find NN architectures suited for deployment at target hardware. These methods incorporate deployment constraints of hardware or latency in their search. Then, they find and train efficient NNs that meet the constraints. However, these NAS methods only satisfy a single deployment target. They need to run repeatedly for each deployment target that doesn't scale well.

\noindent \textbf{Once-For-All Training.} Once-for-all training is a scalable NAS method that satisfies multiple deployment targets. It co-trains models (subnets) that vary in shape and size embedded inside a single supernet (weight-shared). NAS is performed later by extracting specialized subnets from the trained supernet for target hardware. Some of the proposed once-for-all training methods are \ofaCite, \bignasCite, and CompOFA \cite{compofa}. OFA performs Progressive Shrinking (PS) for once-for-all training that trains the full model first and then progressively introduces smaller subnets into the training by dividing the training procedure into multiple training jobs (phases). Compared to OFA, \proposedTraining performs once-for-all training as a single training job and starts shrinking from a partially trained full model to reduce the training cost. BigNAS starts the process of shrinking early and samples multiple subnets at every minibatch. In contrast, \proposedTraining initially only trains the full model and delays the shrinking. Finally, CompOFA changes the architecture search space of OFA and  performs Progressive Shrinking with reduced phases. \proposedTraining algorithmically changes the shrinking procedure in once-for-all training and is complementary to architecture space changes proposed in CompOFA.

\section{Proposed Approach}
\label{sec:proposed}


\begin{figure*}[tb]
 \centering
\begin{subfigure}[b]{0.3\textwidth}
     \includegraphics[width=\columnwidth]{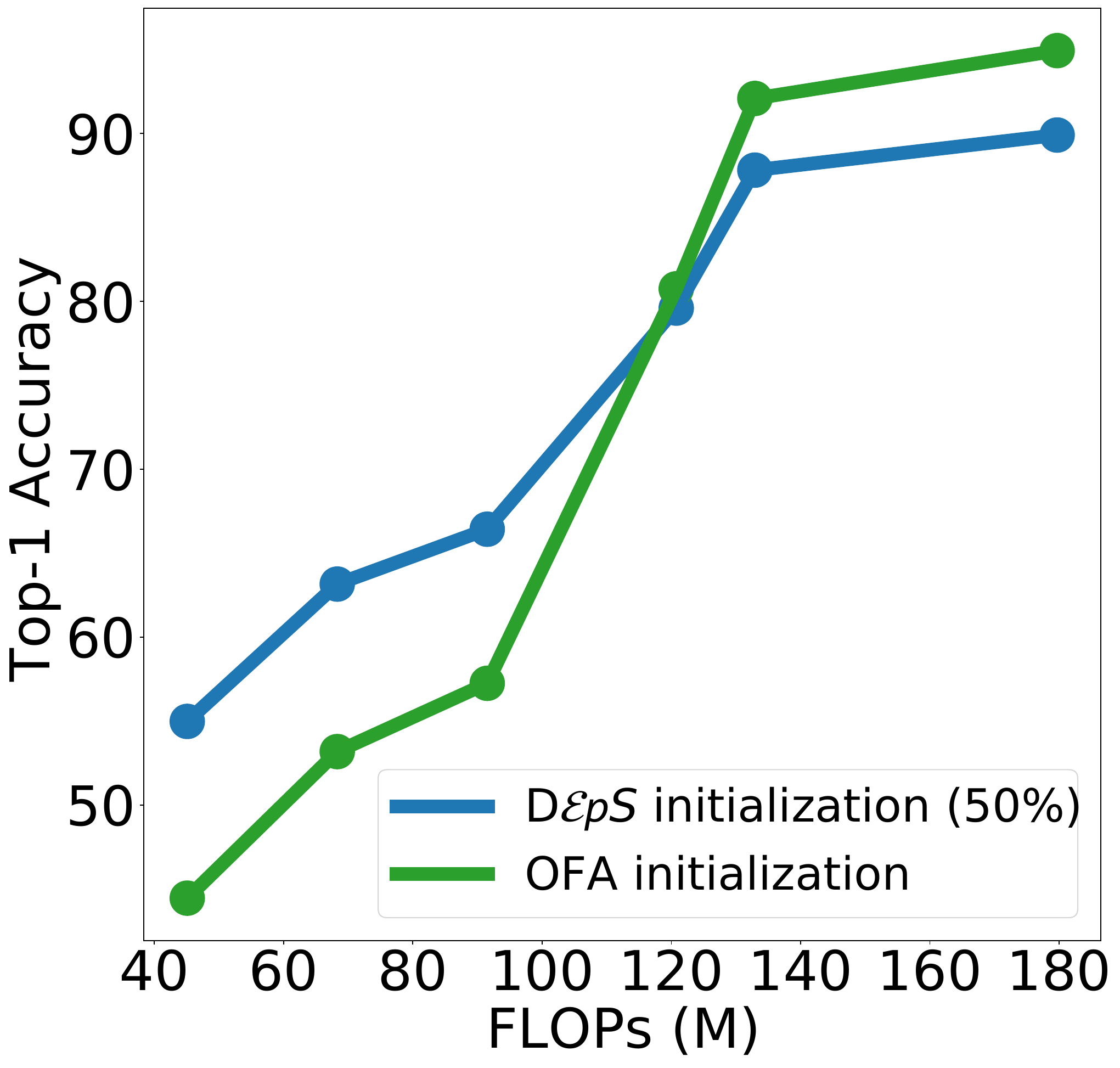}
        \caption{\small CIFAR-10}
             \label{fig:result:rebuttal:fig2_cifar10}
\end{subfigure}
\hfill
\begin{subfigure}[b]{0.3\textwidth}
 \includegraphics[width=\columnwidth]{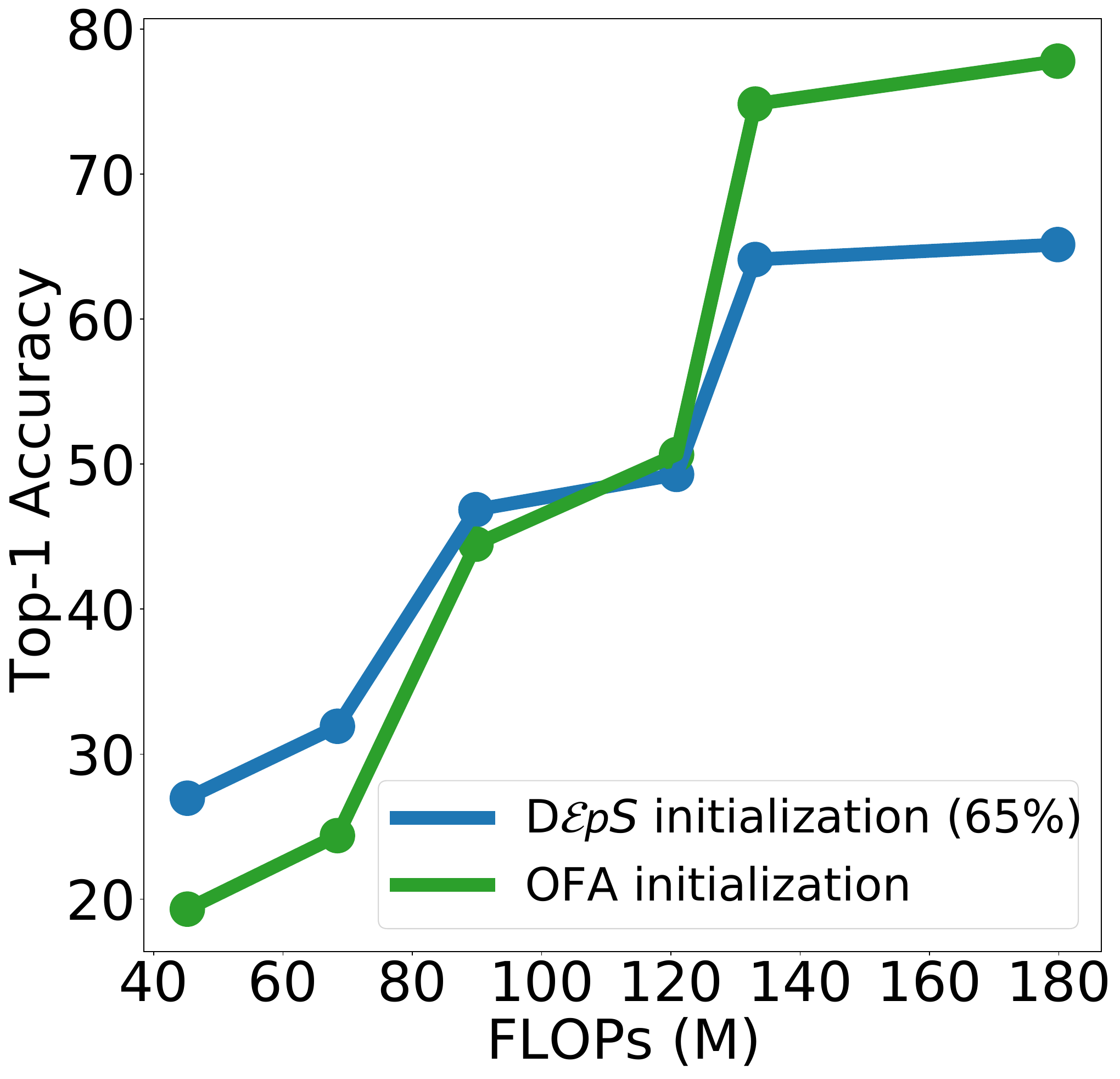}
    \caption{\small CIFAR-100}
         \label{fig:result:rebuttal:fig2_cifar100}
\end{subfigure}
\hfill
\begin{subfigure}[b]{0.3\textwidth}
     \includegraphics[width=\columnwidth]{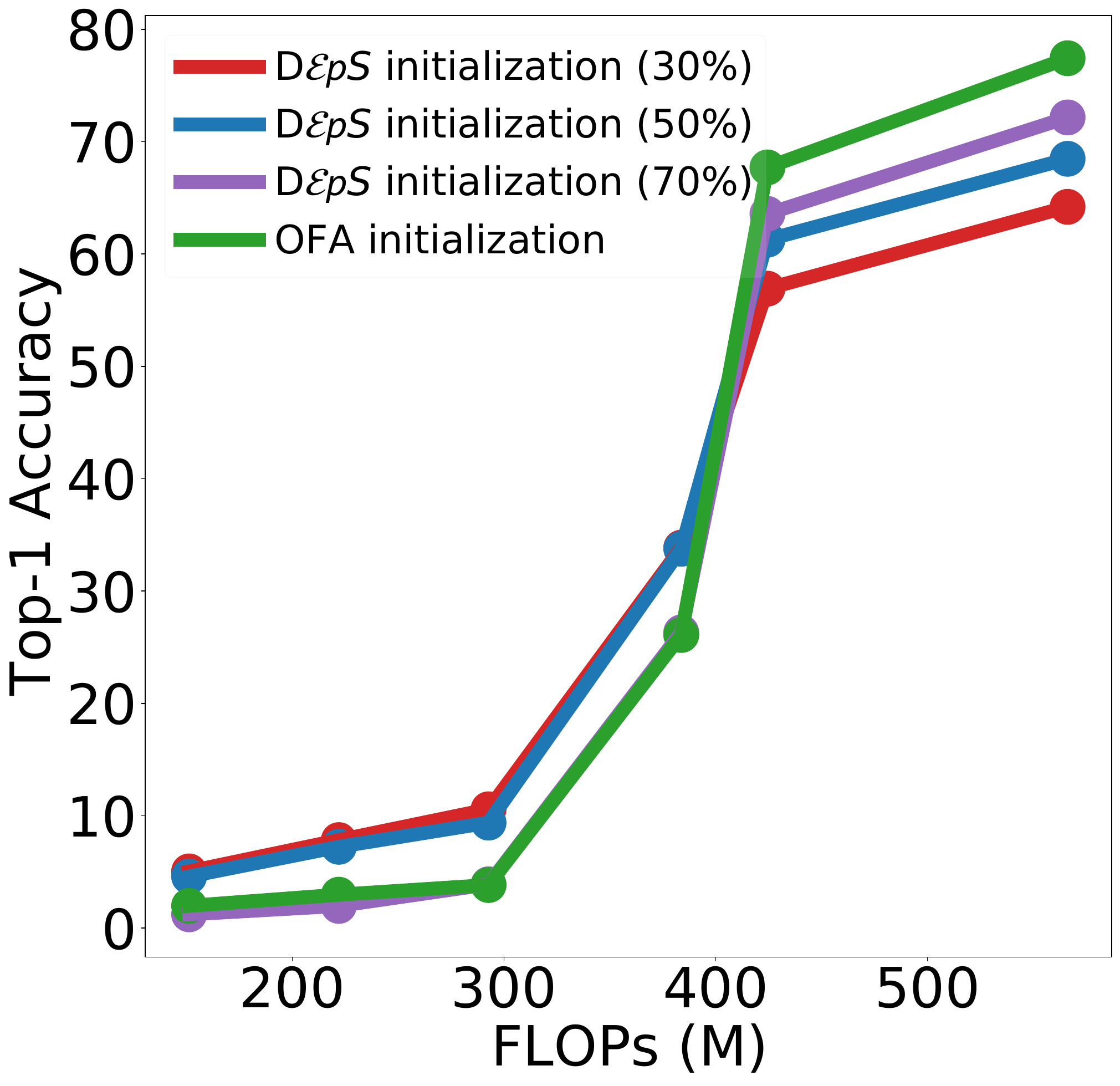}
        \caption{\small ImageNet-1k}
             \label{fig:result:rebuttal:fig2_in1k}
\end{subfigure}

\caption{\small \textbf{Supernetwork initialization.} \proposedTraining provides better initialization for the supernetwork for smaller subnets compared to OFA due to FMWarmup. This validates the hypothesis that the supernet weights become specialized if the full model is trained to completion (OFA), resulting in poorer accuracy of subnetworks with increased training of the full model.}
\label{fig:proposed:init}
\vspace{-0.2in}

\end{figure*}



We present \proposedTraining, a once-for-all training technique that trains supernets in less training time. \proposedTraining consists of three key components that meet the challenges \textbf{C1} and \textbf{C2}. We describe each component in detail and highlight the core contributions of our work. 
\subsection{Full-Model Warmup Period (\fullModelWarmupSym): When to Shrink the Full Model?}
\label{sec:proposed:fm_warmup}
Shrinking the full model at an appropriate time is vital for reducing training cost (meet \textbf{C1}).  Both early or late shrinking isn't sufficient to meet the challenges in once-for-all training.   Early shrinking (\bignasCite in \figref{fig:comparison_sota_dss}) doesn't meet the challenge \textbf{C1}. 
It increases the overall training time as multiple subnets are sampled in each update (increasing per-epoch time) to optimize objective \eqref{eq:problem_formulation}. Early shrinking also requires a lot of hyper-parameter tuning to meet challenge \textbf{C2}. 
It becomes sensitive to training hyper-parameters due to interference. For instance, training the full model with early shrinking  becomes unstable with the standard initialization of the full model \cite{bignas}.

On the other hand, if shrinking happens late after the full model is completely trained (\ofaCite in \figref{fig:comparison_sota_dss}), the supernet weights become too specialized for the full model architecture and require a large number of training epochs to reduce interference. Hence, late shrinking meets challenge \textbf{C2} but not \textbf{C1}. 
\\
We argue that shrinking should occur after the full model is partially trained (warmed up, trained at least $50\%$, proposed approach in \figref{fig:comparison_sota_dss}). 

Delayed Shrinking has numerous advantages. It reduces the overall training time to meet challenge \textbf{C1}. The initial updates in \proposedTraining are cheap compared to early shrinking as only the full model gets trained and no subnets are sampled. Moreover, since supernet weights are not specialized for the full model, \proposedTraining can meet challenge \textbf{C2} in less number of epochs. To validate our hypothesis, we ask whether a partially trained full model serves as a good initialization for the supernet. To do this, we compare the accuracy of small subnets (shrinking) on multiple datasets (CIFAR-10, CIFAR-100, ImageNet-1k) in a mobilenet-based supernet \cite{ofa} when initialized with a partially trained (50\%), and completely trained full model ($\sim$600 MFLOPs) in \figref{fig:proposed:init}.  

\noindent The takeaway from the experiment in \figref{fig:proposed:init} is that a partially-trained full model-based initialization performs better for smaller subnets than the initialization with the full model completely trained. This validates our hypothesis that supernet weights become too specialized if the full model is trained to completion. Hence, warming up the full model helps in meeting challenge \textbf{C1}. We introduce a hyperparameter \fullModelWarmupSym in \proposedTraining that denotes the percentage of total epochs that were used to warmup the full model. \fullModelWarmupSym is usually kept $\geq 50\%$ in \proposedTraining.

\subsection{$\mathcal{E}$-Shrinking: Learning Rates for Subnets}

\label{sec:proposed:eps_shrinking}
In addition to the full model warmup, we propose \LRadjust that enables the full model to reach comparable accuracy with SOTA and meet challenge \textbf{C2a}. \LRadjust ensures that the full model's accuracy doesn't get affected when shrinking is introduced in between its training.
When the shrinking starts, the learning rate of subnets is gradually ramped to reach the full model's learning rate (\LRadjust) as the full model gets sampled with other subnets in each update step. 

Without the gradual warmup, the full model becomes prone to an accuracy drop as the supernet weights change rapidly at the start of shrinking. To understand this change, we compare the updates in the supernet with and without shrinking for a minibatch $\mathcal{B}$. Consider supernet weights $W_t$ at iteration $t$. Without shrinking, the update is given by -

\begin{equation}\label{eq:wo_shrinking}
\small
W^{\text{\textit{\tiny noShrink}}}_{t+1} = W_t - \eta_t \underbrace{\nabla l_{\mathcal{B}}(S(W_t, a_{full})}_{=G^{\mathcal{B},t}_{\text{\textit{\tiny noShrink}}}})
\end{equation}
where $l_{\mathcal{B}}(S(W_t, a_{full})$ denotes the loss of the full model on minibatch $\mathcal{B}$ and equals $ \frac{1}{|\mathcal{B}|} \underset{{x \in \mathcal{B}}}{\sum} l(x , S(W_t, a_{full}))$; $x$ denotes the samples in $\mathcal{B}$. $\eta_t$ denotes the learning rate at iteration $t$ used to update the weights. Whereas introducing shrinking for the same supernet weights $W_t$ yields the following update -

\begin{equation}\label{eq:w_shrinking}
\small
W^{\text{\textit{\tiny Shrink}}}_{t+1} = W_t - \eta_t \underbrace{\left( \overbrace{\underset{a \in \mathcal{U}_k (\mathcal{A})}{\sum}   \nabla l_{\mathcal{B}}\left(S(W_t, a)\right)}^{\text{shrinking}} \right)}_{=G^{\mathcal{B},t}_{\text{\textit{\tiny Shrink}}}}
\end{equation}
where $\small{\mathcal{U}_k (A)}$ denotes uniformly sampling $k$ subnets from the architecture space $\mathcal{A}$.  This update step is the approximation of the objective \eqref{eq:problem_formulation}. Clearly, the updates differ, it is \textit{improbable} that  $W^{\text{\textit{\tiny Shrink}}}_{t+1} = W^{\text{\textit{\tiny NoShrink}}}_{t+1}$. This difference in updates causes the supernet weights to change rapidly when shrinking is introduced. The rapid change in supernet weights causes degradation in the full model's accuracy. To avoid rapid changes in weights, a widely adopted technique is to use less aggressive learning rates via learning rate warmup schedules \cite{lr_warmup_2, lr_warmup_1}. 

However, applying such principles in the context of weight-sharing is non-trivial but at the same time important.
Our key idea is two-fold to a) always sample the full model with other subnets while shrinking, and b) use less aggressive learning rates for subnets at the start of shrinking. 
Particularly, it is important to ensure $G^{\mathcal{B},t}_{\text{\textit{\scriptsize noShrink}}} \approx G^{\mathcal{B},t}_{\text{\textit{\scriptsize Shrink}}}$ to make $W^{\text{\textit{\tiny Shrink}}}_{t+1} \approx W^{\text{\textit{\tiny NoShrink}}}_{t+1}$ initially when the shrinking starts. 
To do this, we introduce a parameter $\mathcal{E}$ that controls the effective learning rate of subnets and makes $G^{\mathcal{B},t}_{\text{\textit{\scriptsize noShrink}}} \approx G^{\mathcal{B},t}_{\text{\textit{\scriptsize Shrink}}}$. The gradient in \LRadjust is given as follows -
\begin{equation}\label{eq:ep_shrinking}
\small
G^{\mathcal{B},t}_{\text{\textit{\tiny Shrink}}}(\mathcal{E}_t) = G^{\mathcal{B},t}_{\text{\textit{\scriptsize noShrink}}} + \overbrace{\mathcal{E}_t*\underset{a \in \mathcal{U}_{k-1} (\mathcal{A} \setminus \left\{ a_{full}\right\})}{\sum}   \nabla l_{\mathcal{B}}\left(S(W_t, a)\right)}^{\mathcal{E}-\text{shrinking}}
\end{equation}
where $\mathcal{E}_t \in (0,1]$.  Note that the effective learning rate becomes $\eta_t*\mathcal{E}_t$ for subnets and remains $\eta_t$ for the full model in \LRadjust. Hence, slowly increasing $\mathcal{E}_t$ warms up the effective learning of subnets. We start with a small value of $\mathcal{E}_t$ (=$10^{-4}$) and increment it by a constant amount to reach $1$. Once $\mathcal{E}_t$ reaches $1$, it stays constant for the rest of the training. 
We empirically verify if $G^{\mathcal{B},t}_{\text{\textit{\scriptsize noShrink}}}$, $ G^{\mathcal{B},t}_{\text{\textit{\scriptsize Shrink}}}$ differ in magnitude ($l_2$-norm) and direction (cosine similarity) and
\begin{figure}[tb]
 \centering
  \begin{subfigure}[h]{0.44\columnwidth}
         \includegraphics[width=\columnwidth]{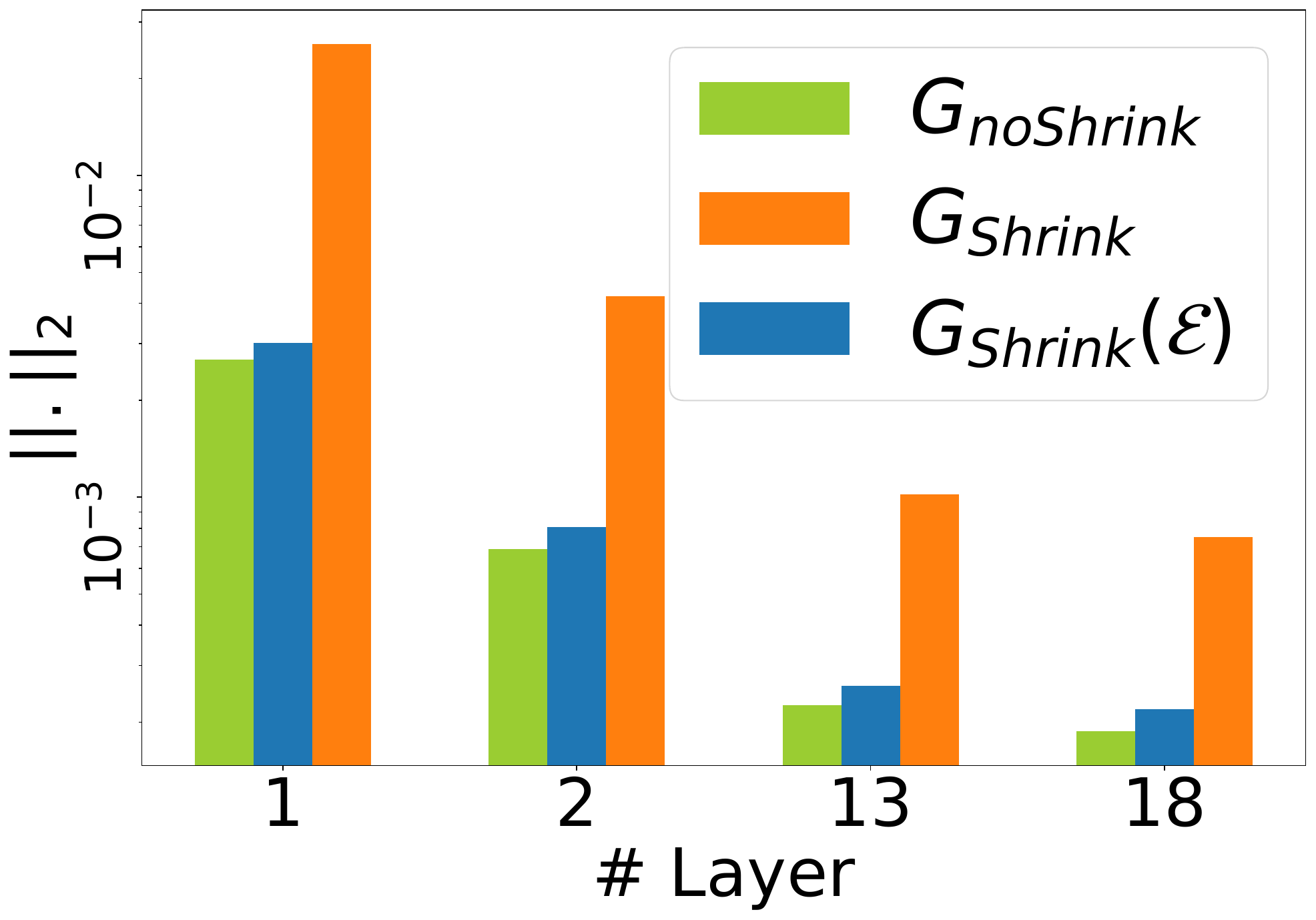}
        	\caption{\small Magnitude ($||.||_2$)}
                 \label{fig:prop:eps_shrink:grad_magnitude}
         \end{subfigure}
        \begin{subfigure}[h]{0.44\columnwidth}
         \includegraphics[width=\columnwidth]{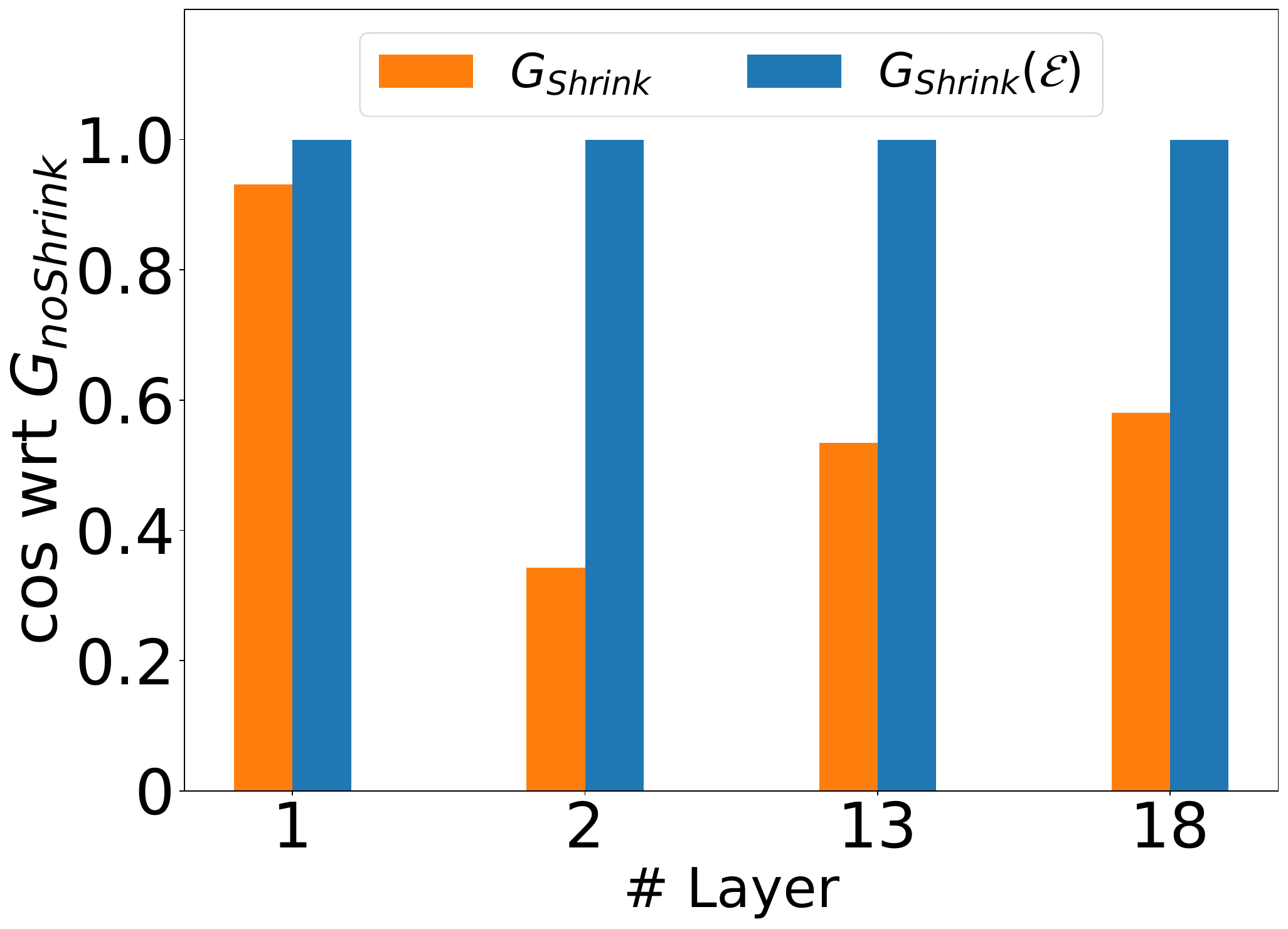}
        	\caption{\small Direction (cos. sim.)}
                 \label{fig:prop:eps_shrink:grad_direction}
         \end{subfigure}
    \caption{\small \textbf{Gradients w/ \& w/o Shrinking on Mobilenet-Based Supernet.} Delayed Shrinking causes gradients ($G_{\text{\textit{Shrink}}}$) to differ from the full model gradient ($G_{\text{\textit{noShrink}}}$) leading to rapid changes in the supernet's weights. \LRadjust's gradient ($G_{\text{\textit{Shrink}}}(\mathcal{E})$) reduces such differences and avoids rapid weight changes.
    }
    \label{fig:prop:eps_shrink}
    \vspace{-0.2in}

\end{figure}

whether \LRadjust is able to reduce the differences with  $G^{\mathcal{B},t}_{\text{\textit{\tiny noShrink}}}(\mathcal{E}_t)$. \figref{fig:prop:eps_shrink} compares the magnitude and direction of the gradients of the full model ($G_{\text{\textit{\scriptsize noShrink}}}$),  shrinking ($G_{\text{\textit{\scriptsize Shrink}}}$) and \LRadjust ($G_{\text{\textit{\tiny noShrink}}}(\mathcal{E})$) ($\mathcal{E}=0.001$) on the weights of a mobilenet-based supernet \cite{ofa} for the ImageNet dataset 
\cite{imagenet}. $G_{\text{\textit{\scriptsize noShrink}}}$ and $G_{\text{\textit{\scriptsize Shrink}}}$ differ both in magnitude and direction across supernet layers. 
\begin{wrapfigure}{b}{0.5\columnwidth}
\vspace{-0.2in}
     \includegraphics[width=0.45\columnwidth]{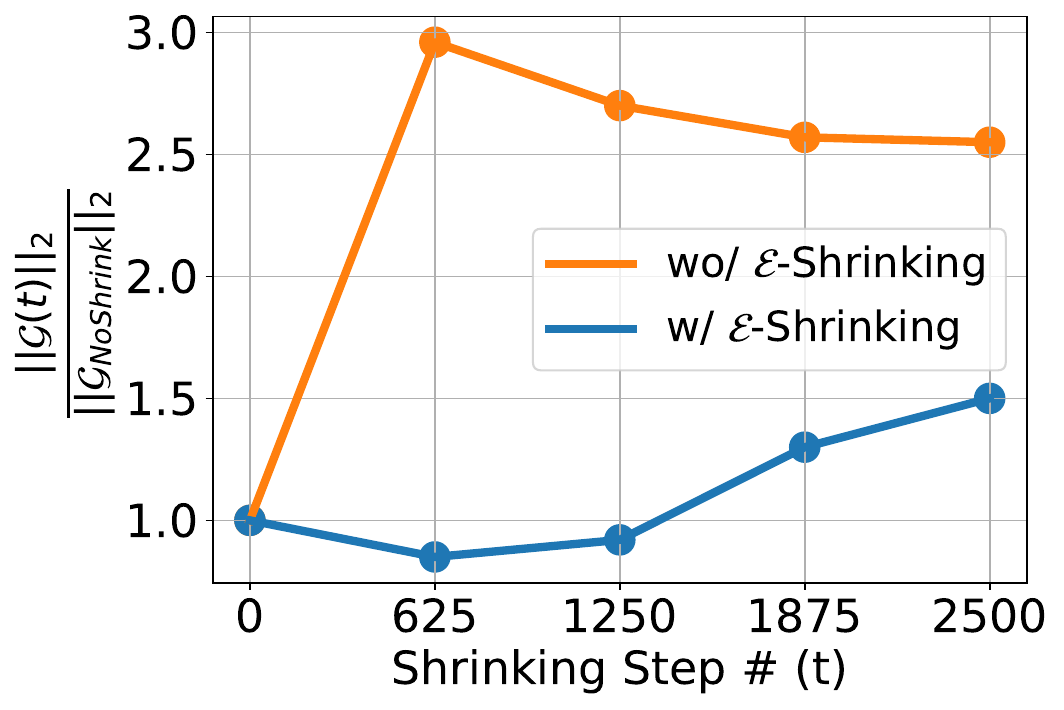}
        \caption{\small 
            \textbf{Gradient Magnitude Over Time.} Gradient magnitude with ($\mathcal{G}_{shrink}(\mathcal{E}, t)$) and without ($\mathcal{G}_{shrink}(t)$) $\mathcal{E}$-shrinking is compared w.r.t the initial full model gradient ($\mathcal{G}_{Noshrink}$) over shrinking steps. $\mathcal{E}$-shrinking avoids sudden changes in the supernet parameters by lowering the gradient magnitude.}
     \label{fig:proposed:gradient_mag_time}
    \vspace{-0.5in}
\end{wrapfigure}
The magnitude of   $G_{\text{\textit{\scriptsize Shrink}}}$ is an order of magnitude higher than $G_{\text{\textit{\scriptsize noShrink}}}$ for early layers.
\LRadjust maintains the low magnitude of gradient throughout the training as shown in \figref{fig:proposed:gradient_mag_time}. The magnitude of $G_{\text{\textit{\scriptsize Shrink}}}$ is consistently higher than $G_{\text{\textit{\scriptsize Shrink}}}(\mathcal{E}_t)$ when normalized with the magnitude of $G_{\text{\textit{\scriptsize noShrink}}}$. Such differences cause poor convergence at the start of shrinking and often lead to accuracy drops. 
Whereas, $G_{\text{\textit{\tiny noShrink}}}(\mathcal{E})$ has minimal differences \wrt $G_{\text{\textit{\scriptsize noShrink}}}$ enabling healthy convergence and no potential accuracy drops.

\subsection{\proposedinplaceKD : In-Place Knowledge Distillation (KD) from Warmed-up Full Model}\label{sec:ikd_warmup}
We now discuss \proposedinplaceKD that distills knowledge from the full model to subnets and meets challenge \textbf{C2b}. Effectively distilling the knowledge from the full model becomes non-trivial due to weight-sharing. On one hand, KD requires the supernet weights biased to the full model to offer meaningful knowledge transfer to subnets. On the other hand, having a large bias in the supernet weights toward the full model may result in subnets' sub-optimal performance since the weights are shared. 
To tackle this trade-off, \ofaCite biases the supernet weights to a trained full model and then uses it to perform vanilla-KD \cite{vanillakd}. However, this results in a long training time during shrinking as the supernet weights are trained to fit subnets' architectures. Another approach like \bignasCite doesn't bias the shared weights to the full model by using inplace-KD \cite{inplacekd} but lacks in providing rich knowledge transfer to subnets (initially).

\noindent This is because inplace-KD distills the knowledge "on the fly" to other subnets as the full model gets trained from randomly initialized weights. Precisely, the full model predictions become ground truth for other subnets. Hence, when the full model is under-trained initially, it doesn't offer rich knowledge transfer.
\\
We believe that the proposed delayed shrinking has an added advantage \wrt KD for once-for-all training --- the partially trained full model ($50/60\%$ trained) is rich enough to provide meaningful knowledge transfer to the subnets and doesn't bias the supernet weights to the full model. It has been shown that for vanilla-KD \cite{vanillakd}, partially trained (intermediate) models provide a comparable or at times better knowledge transfer than the completely trained models
\cite{poorTeacherKD, partialKD}. This is because they provide more information about non-target classes than the trained models  \cite{poorTeacherKD}. We use this insight in \proposedTraining that performs inplace-KD from a partially trained full model (\proposedinplaceKD). 

\noindent \proposedinplaceKD offers two advantages, it --- a) distills knowledge from multiple progressively better partially trained models as the full model gets trained (unlike a single partially/fully trained model used in vanilla-KD \cite{partialKD}), and b) provides rich knowledge transfer to the subnets at all times (unlike inplace-KD \cite{inplacekd} that uses under-trained full model initially).

\section{Experiments}
\label{sec:experiments}
We establish that \proposedTraining \textbf{a}) reduces training cost \wrt SOTA in once-for-all training \cite{ofa, compofa, bignas}, \textbf{b}) performs at-par or better than SOTA's accuracy across subnets (covering the entire range of architectural space), \textbf{c}) generalizes across datasets, \textbf{d}) generalizes to different deep neural network (DNN) architecture spaces, and \textbf{e}) produces specialized subnets for target hardware without retraining (once-for-all property). We also aim to demonstrate attribution of benefits in \proposedTraining by providing detailed ablation on \textbf{a}) a full model warmup period: empirically demonstrating a sweet spot, \textbf{b}) \LRadjust: showing healthy convergence, and \textbf{c}) \proposedinplaceKD : distilling knowledge better than existing distillation approaches in weight-sharing.

\subsection{Setup}

\noindent\textbf{Baselines.} We first compare \proposedTraining with the other NAS methods or efficient DNNs \cite{fbnet_v2, mbv3, efficientnet, proxylessnas} \wrt accuracy. Then, We compare \proposedTraining with once-for-all training techniques --- \ofaCite, \bignasCite, CompOFA \cite{compofa} \wrt both training cost and accuracy of subnets spanned across supernet's FLOP range. The training time of all the techniques is measured on NVIDIA A40 GPUs. As once-for-all training trains multiple subnets, the comparison is done by uniformly dividing the entire FLOP range into 6 buckets and picking the most accurate subnet from each bucket for every baseline. 

\noindent\textbf{Success Metrics.} \proposedTraining is compared against the baselines on the following success metrics --- a) Training cost measured in GPU hours or dollars (lower is better), b) \textit{Pareto-frontier}: Accuracy of best-performing subnets as a function of FLOPs/latency. To compare Pareto-frontiers obtained from different baselines, we use a metric called \textit{mean pareto accuracy} that is defined as the area under the curve (AUC) of accuracy and normalized FLOPs/latency. The higher the mean pareto accuracy the better.

\noindent\textbf{Datasets.} We evaluate all methods on CIFAR10/100 \cite{cifar}, ImageNet-100 \cite{imagenet100} and ImageNet-1k \cite{imagenet1k} datasets. The complexity of datasets progressively increases from CIFAR10 to ImageNet-1k. The datasets vary in the number of classes, image resolution, and number of train/test samples.

\noindent\textbf{DNN Architecture Space.} All methods are trained on the supernets derived from two different DNN architecture spaces --- MobilenetV3 \cite{mbv3} and ProxylessNAS \cite{proxylessnas} (same as \ofaCite). The base architecture of ProxylessNAS is derived from ProxylessNAS run for the GPU as a target device. To avoid confounding, we evaluate all baselines on the same DNN architecture space. 

\noindent\textbf{Training Hyper-parameters.}
The training hyper-parameters of \proposedTraining are similar to the hyper-params of the full model training. The  hyper-parameters for MobilenetV3, and ProxylessNAS training are borrowed from \cite{mbv3} and \cite{proxylessnas} respectively. Specifically, we use SGD with Nesterov momentum 0.9, a CosineAnnealing LR \cite{sgdr} schedule, and weight decay $3e^{-5}$. Unless specified, the shrinking is introduced in \proposedTraining after the full model gets $\sim 50\%$ trained.

\begin{table}[tb]
\centering
\footnotesize
\renewcommand{\arraystretch}{1}
\resizebox{0.9\textwidth}{!}{
\begin{tabular}{|c|c|c|c|}
\hline
Group & Approach & MACs (M) & Top-1 Test Acc (\%) \\ \hline
\multirow{2}{*}{0-100 (M)} & 
 OFA \cite{ofa} & 67 & 70.5 \\
 & \textbf{\proposedTraining} & \textbf{67} & \textbf{72.3} \\
\hline

\multirow{2}{*}{100-200 (M)} & 
 OFA \cite{ofa} & 141 & 71.6 \\
 & \textbf{\proposedTraining} & \textbf{141} & \textbf{73.7} \\
\hline

\multirow{4}{*}{200-300 (M)} & 
FBNetv2 \cite{fbnet_v2} & 238 & 76.0 \\
 & BigNAS \cite{bignas} & 242 & 76.5 \\
 & OFA \cite{ofa} & 230 & 76 \\
 & \textbf{\proposedTraining} & \textbf{230} & \textbf{77.3} \\
\hline

\multirow{5}{*}{300-400 (M)} &
MNasnet \cite{mnasnet} & 315 & 75.2 \\
 & ProxylessNAS \cite{proxylessnas} & 320 & 74.6 \\
 & FBNetv2 \cite{fbnet_v2} & 325 & 77.2 \\
 & MobileNetV3 \cite{mbv3} & 356 & 76.6 \\
 & EfficientNetB0 \cite{efficientnet} & 390 & 77.3 \\
\hline
\end{tabular}
}
\caption{\small Comparison of \proposedTraining with state of the art neural architecture search approaches on Imagenet. \proposedTraining consistently outperforms the baselines.}
\label{tab:macs_performance}
\vspace{-0.2in}
\end{table}


\subsection{Evaluation}
\noindent \textbf{Comparison with NAS methods/Efficient Nets on ImageNet.} We compare \proposedTraining with MobilenetV3 \cite{mbv3}, FBNet \cite{fbnet_v2}, ProxylessNAS \cite{proxylessnas}, BigNAS \cite{bignas} and efficient nets \cite{efficientnet} on the Imagenet Dataset.

\noindent \textit{Takeaway.} \tableref{tab:macs_performance} compares accuracy vs MACs of the baselines. \proposedTraining consistently surpasses the baselines over multiple MAC ranges. Especially in the lower MAC region (0-100M), \proposedTraining is \textbf{1.8\%} more accurate. Moreover, in the larger MAC region (200-300M), \proposedTraining achieves $77.3\%$ accuracy with upto $1.69$x MACs improvement compared to the baselines (efficientNet-B0). \proposedTraining benefits from supernet initialization and effective knowledge distillation to get superior performance. 

\noindent \textbf{Comparison with Once-for-all training methods on ImageNet}
We now demonstrate the accuracy and training cost benefits of \proposedTraining on ImageNet dataset \cite{imagenet}. \tableref{tab:cost_compare_sota} compares \proposedTraining with the baselines\footnote{\label{footnote:compofa} \small There is no available open-source checkpoint of CompOFA \cite{compofa}. CompOFA claims to match OFA's Pareto-optimality. Hence, we report Pareto-frontier of \ofaCite instead.} on a) the upper-bound (largest subnet) and lower-bound (smallest subnet) top1 accuracy, and b) GPU hours and dollar costs.

\noindent \textit{Takeaway.}
\proposedTraining is atleast \textbf{2$\%$} more accurate at $150$ MACs (smallest subnet) than baselines and at-par \wrt accuracy at $230$ MACs (largest subnet). \proposedTraining matches the Pareto-optimality of baselines (with highest mean pareto accuracy) at a reduced training cost (least among all the baselines). It takes $1.8$x and $2.5$x less dollar cost (or GPU hours) than OFA and BigNAS respectively.
\\
\noindent The training cost improvement of \proposedTraining comes due to \fullModelWarmup. \fullModelWarmup allows \proposedTraining to train subnets in less number of total epochs (lowest among the baselines) and a lower average time per epoch than BigNAS (\tableref{tab:cost_compare_sota}). The full model's accuracy (largest subnet in \tableref{tab:cost_compare_sota}) is improved as \LRadjust enables its smooth convergence. Finally, \proposedTraining improves accuracy at lower FLOPs ($150$ MACs) as \proposedinplaceKD distills knowledge effectively in once-for-all training.
\begin{table}[tb]
\large
\centering
\renewcommand{\arraystretch}{1.25}
\setlength{\arrayrulewidth}{0.3mm}

\resizebox{\textwidth}{!}{
\begin{tabular}{|c|cc|cc|p{3cm}|cp{1.7cm}p{1.7cm}c|}
\hline
\multirow{2}{*}{Approach} &
  \multicolumn{2}{c|}{Smallest Subnet} &
  \multicolumn{2}{c|}{Largest Subnet} &
  \multirow{2}{*}{mean pareto acc.} &
  \multicolumn{4}{c|}{Training Cost} \\ \cline{2-5} \cline{7-10} 
 &
  \multicolumn{1}{c|}{Acc(\%)} &
  MACs (M) &
  \multicolumn{1}{c|}{Acc (\%)} &
  \multicolumn{1}{c|}{MACs (M)} &
  &
  \multicolumn{1}{c|}{\# Epochs} &
  \multicolumn{1}{p{1.7cm}|}{Avg. GPU min. / epoch} &
  \multicolumn{1}{p{1.7cm}|}{Total Time (GPU hours.)} &
  Dollar Cost (\$) \\ \hline
\ofaCite &
  \multicolumn{1}{c|}{71.8} &
  150 &
  \multicolumn{1}{c|}{77.2} &
  230 &
  75.77 &
  \multicolumn{1}{c|}{605} &
  \multicolumn{1}{c|}{125} &
  \multicolumn{1}{c|}{1256} &
  2675 \\
CompOFA \cite{compofa} &
  \multicolumn{1}{c|}{-\textsuperscript{\ref{footnote:compofa}}} &
  150 &
  \multicolumn{1}{c|}{-\textsuperscript{\ref{footnote:compofa}}} &
  230 &
  - \textsuperscript{\ref{footnote:compofa}} &
  \multicolumn{1}{c|}{330} &
  \multicolumn{1}{c|}{142} &
  \multicolumn{1}{c|}{782} &
  1665 \\
\bignasCite &
  \multicolumn{1}{c|}{70.6} &
  150 &
  \multicolumn{1}{c|}{74} &
  230 &
  72.51 &
  \multicolumn{1}{c|}{400} &
  \multicolumn{1}{c|}{266} &
  \multicolumn{1}{c|}{1778} &
  3787 \\
\proposedTraining &
  \multicolumn{1}{c|}{\textbf{73.8}} &
  150 &
  \multicolumn{1}{c|}{\textbf{77.3}} &
  230 &
  \textbf{75.81} &
  \multicolumn{1}{c|}{\textbf{270}} &
  \multicolumn{1}{c|}{155} &
  \multicolumn{1}{c|}{\textbf{700}} &
  \textbf{1491} \\ \hline
\end{tabular}
}
\caption{\small \textbf{Comparison of \proposedTraining vs SOTA on ImageNet.} Accuracy and Training Cost comparison of \proposedTraining against SOTA approaches are shown for MobilenetV3-based architecture space. \proposedTraining outperforms SOTA and achieves $2\%$ better accuracy for the smallest subnet and is at-par with the largest subnet (full model) respectively at $1.8$x training cost reduction (in \$) compared to OFA. Dollar-cost is calculated based on the on-demand prices for A40 GPUs from exoscale.com}
\label{tab:cost_compare_sota}
\vspace{-0.3in}
\end{table}
\begin{figure}[t]
 \centering
\begin{subfigure}[b]{0.24\columnwidth}
     \includegraphics[width=\columnwidth]{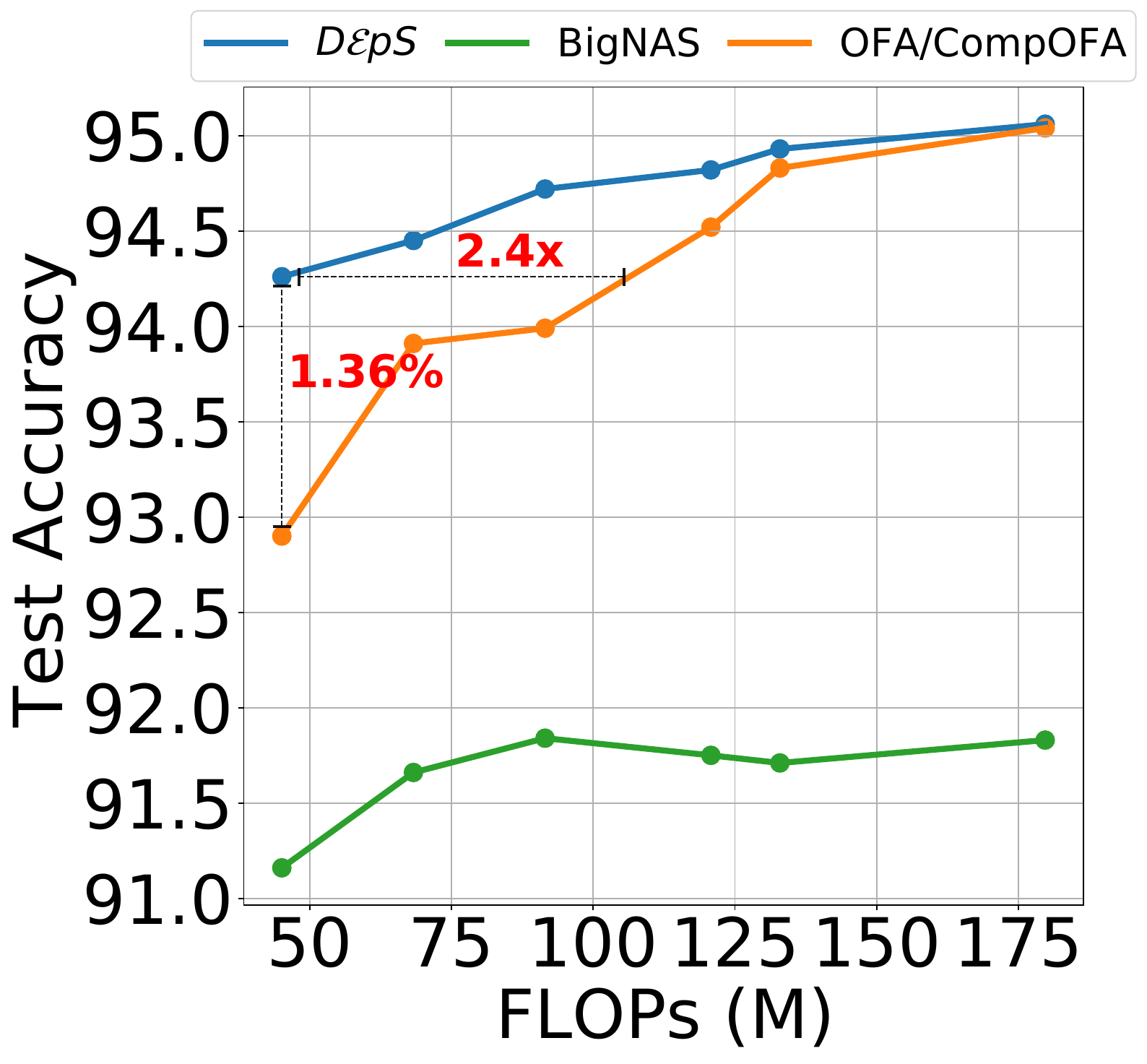}
        \caption{\small CIFAR-10 }
             \label{fig:result:dataset_generalize:cifar_10}
\end{subfigure}
    \hfill
    \begin{subfigure}[b]{0.24\columnwidth}
     \includegraphics[width=\columnwidth]{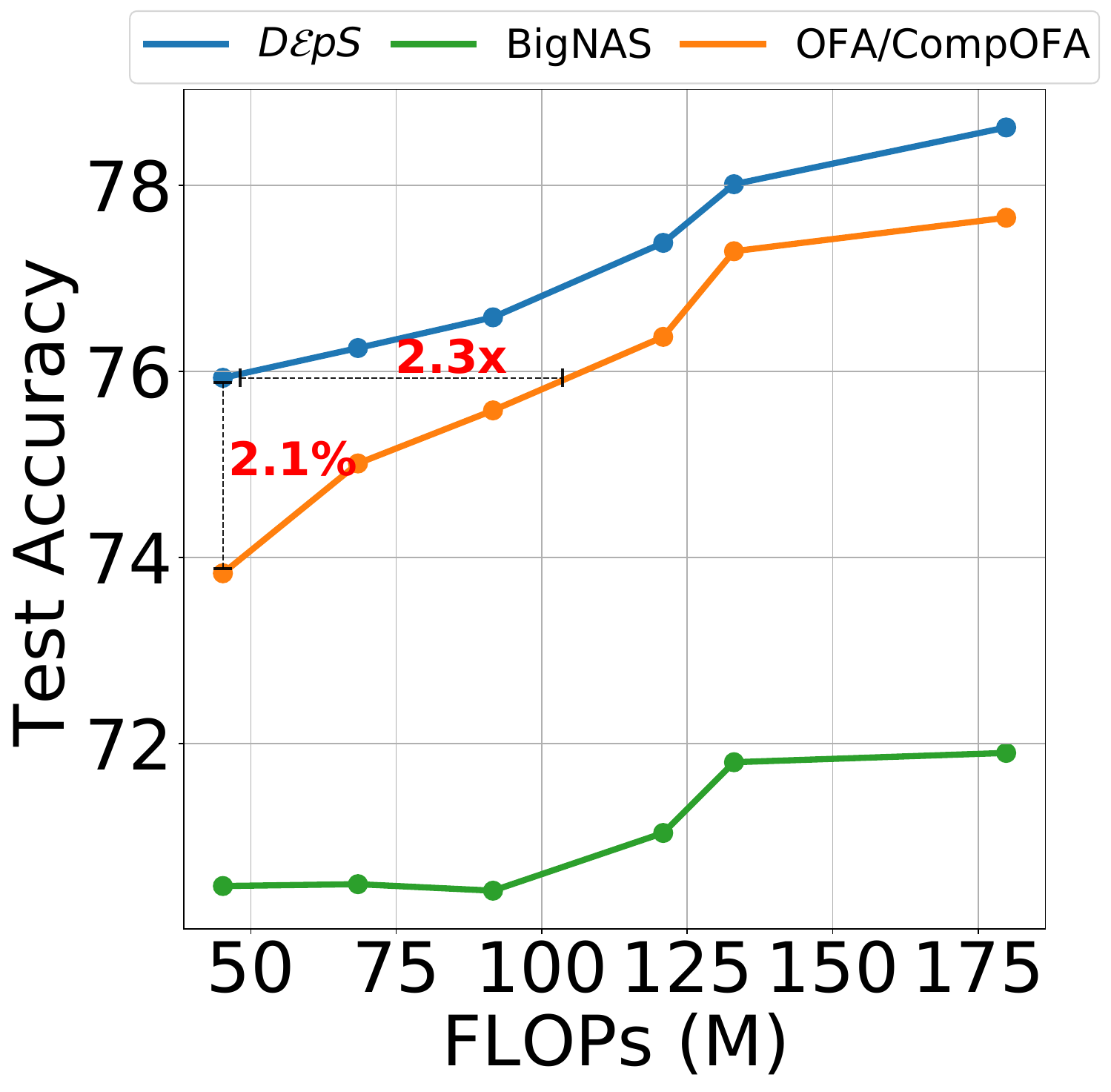}
        \caption{\small CIFAR-100}
             \label{fig:result:dataset_generalize:cifar_100}
\end{subfigure}
     \hfill
\begin{subfigure}[b]{0.24\columnwidth}
    \includegraphics[width=\columnwidth]{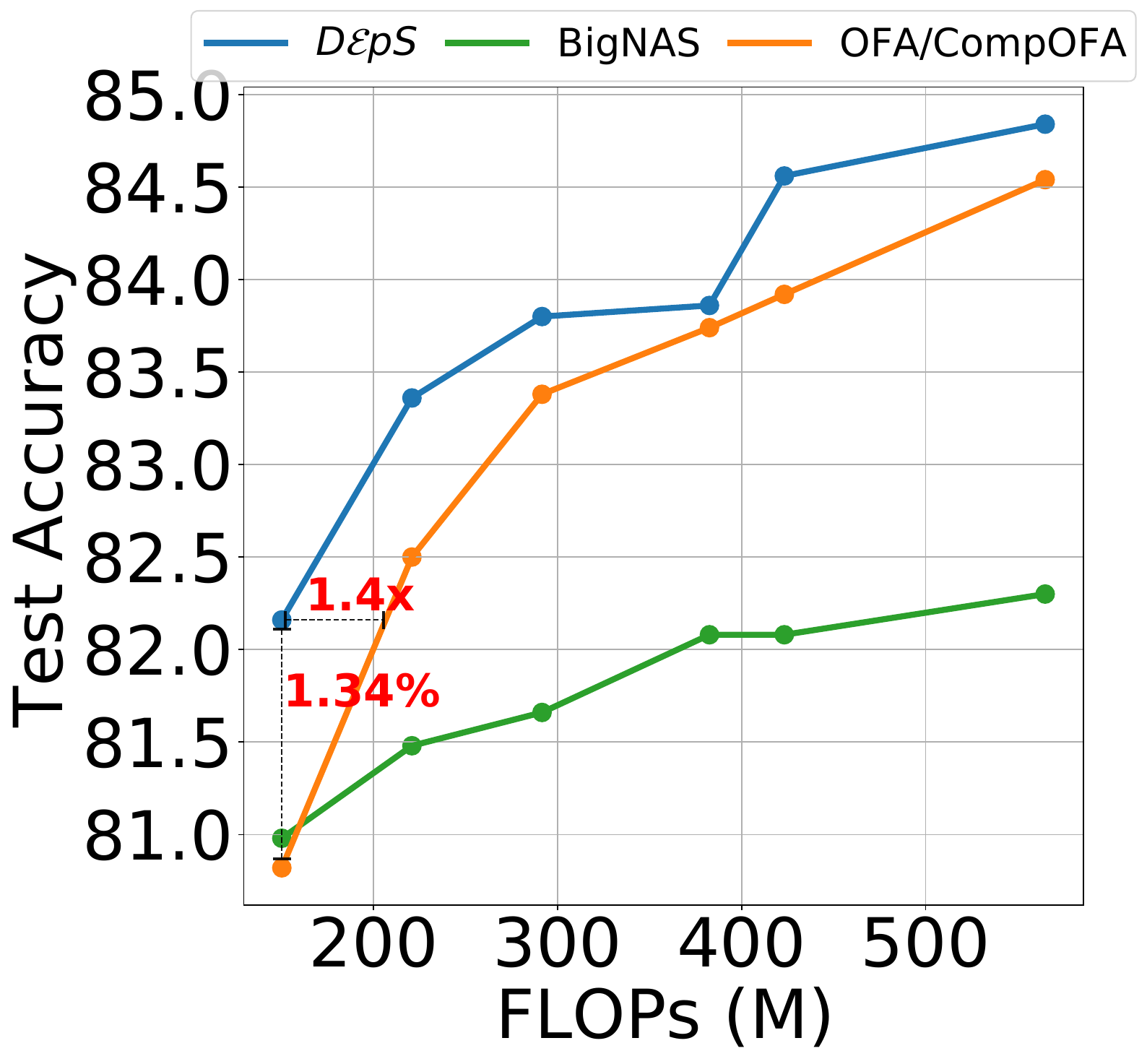}
    \caption{\small ImageNet-100}
         \label{fig:result:dataset_generalize:in_100}
\end{subfigure}
     \hfill
\begin{subfigure}[b]{0.24\columnwidth}
    \includegraphics[width=\columnwidth]{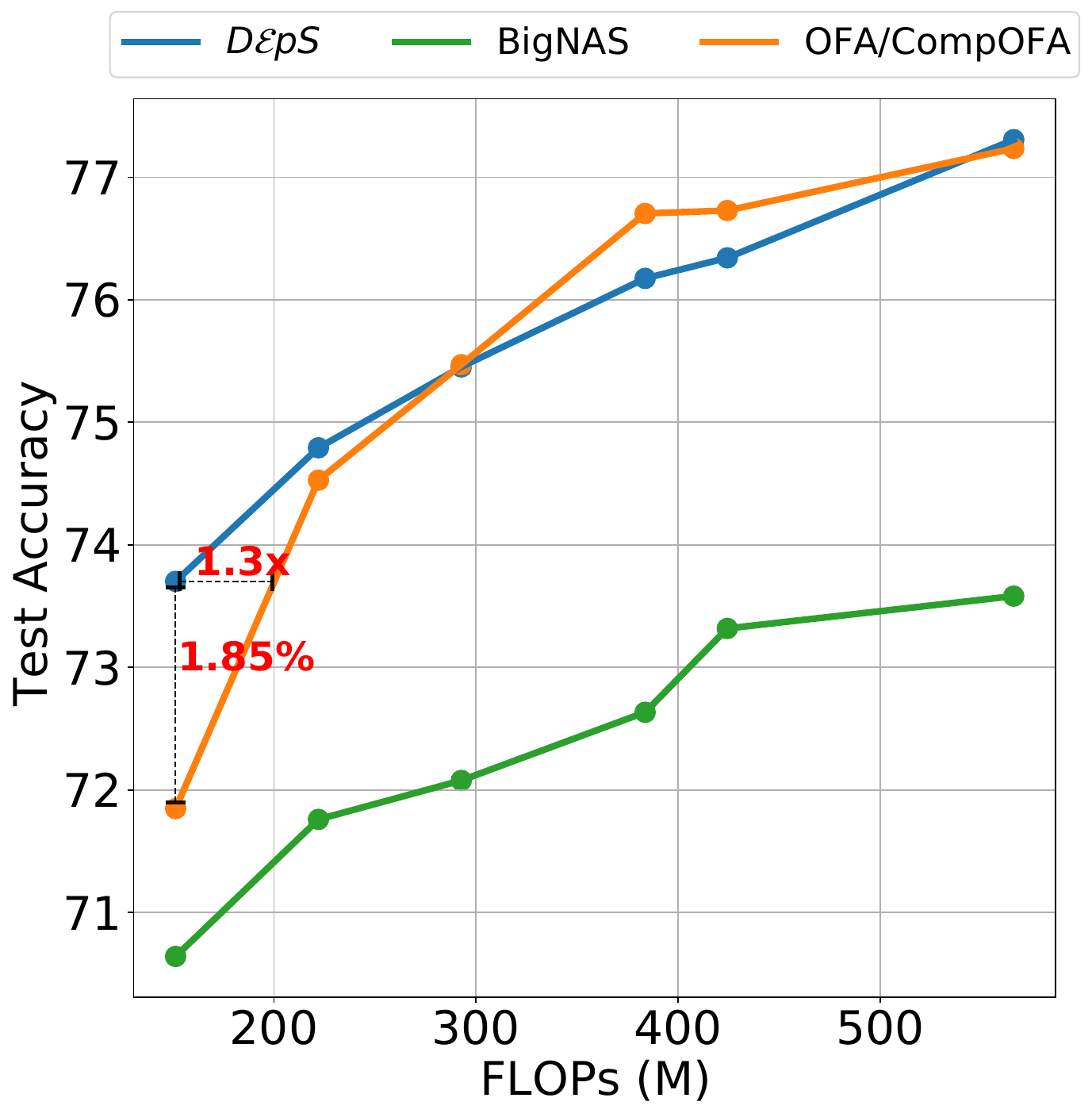}
    \caption{\small ImageNet-1k}
         \label{fig:result:dataset_generalize:in_1k}
\end{subfigure}
\caption{\small \textbf{\proposedTraining's Accuracy Improvement across Datasets.} The comparison of \proposedTraining with the baselines is shown \wrt accuracy (of subnets)  for CIFAR10/100, ImageNet-100, and ImageNet-1k datasets. \proposedTraining consistently outperforms the baselines across all the datasets and achieves upto $2.1\%$ better accuracy for the same FLOPs or upto $2.3$x FLOP reduction at same accuracy.}
\label{fig:result:dataset_generalize}
\vspace{-0.3in}
\end{figure}

\noindent \textbf{Generalization across datasets.}
We establish that the accuracy improvements of \proposedTraining generalize to other vision datasets.  

\noindent \textit{Training Details.} \proposedTraining uses the standard hyper-parameters of the MobileNetV3 for all the datasets using SGD with cosine learning rate decay and nestrov momentum, and shrinking is introduced when the full model is 50\% trained. For OFA, we first train the largest network independently. Shrinking occurs after the full model is completely trained and vanilla KD is used for distillation. The depth and expand phases are run for 100 epochs each. The initial learning rate of different phases is set as per \ofaCite. BigNAS uses RMSProp optimizer with its proposed hyper-parameters for ImageNet-1k. However, we use SGD optimizer in BigNAS for CIFAR10/100 and ImageNet-100 datasets as we empirically find that SGD performs better than RMSProp on these datasets. 
\figref{fig:result:dataset_generalize} compares the Pareto-frontiers of top1 test accuracy and FLOPs obtained from each baseline across various datasets. The subnets are present in six different FLOP buckets that uniformly divide the supernet's FLOP range. The comparison includes the performance of the smallest and largest subnets to measure the lower-bound and upper-bound test accuracy reached by the baselines. 
\\
\textbf{Takeaway.} 
\proposedTraining outperforms baselines \wrt accuracy of smaller subnets ($\leq 300$ MFLOPs) on all the datasets. It achieves slightly better or at-par accuracy for larger subnets ($\geq 300$ MFLOPs) than OFA/CompOFA. \proposedTraining outperforms BigNAS 
and achieves a better Pareto-Frontier across all the datasets.
\begin{wrapfigure}{t}{0.5\columnwidth}
    \vspace{-0.2in}
    \includegraphics[width=0.5\columnwidth]{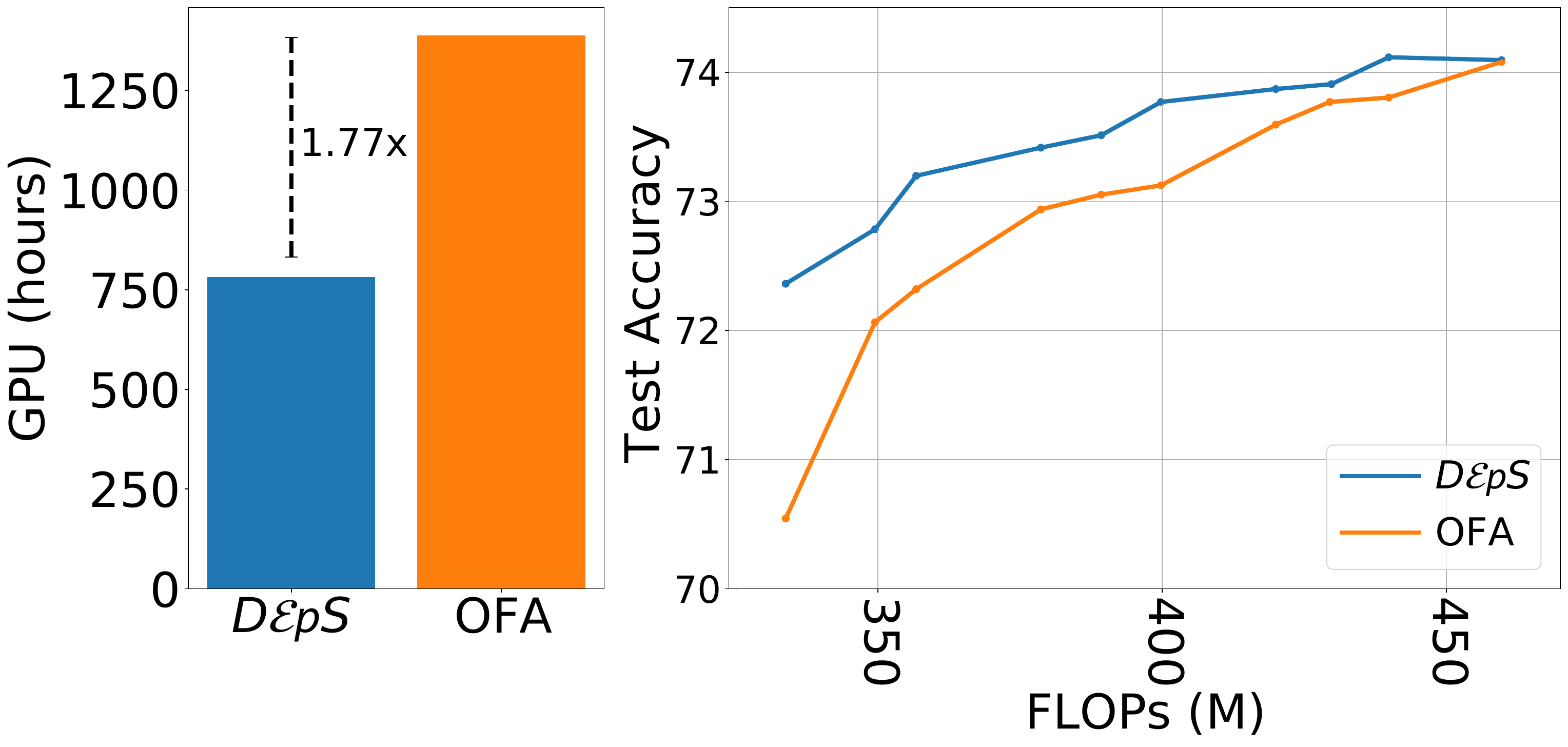}
    \vspace{-0.2in}
    \caption{\small
    \textbf{\proposedTraining on ProxyLessNAS architecture space:} superior Pareto-Frontier with a $1.8\%$ improvement in ImageNet-1k test accuracy on the smallest subnet.}
\label{fig:result:network_generalization:performance}
\vspace{-0.2in}
\end{wrapfigure} 
\noindent \textbf{Generalization across DNN-Architecture Spaces.}
We demonstrate that \proposedTraining generalizes to other DNN-architecture spaces.
We train \proposedTraining on ImageNet-1k dataset using ProxylessNAS-based supernet (DNN-architecture space) with training-hyperparameters borrowed from \cite{proxylessnas}. \figref{fig:result:network_generalization:performance} compares Pareto-frontiers obtained from \proposedTraining and OFA on ImageNet-1k dataset.

\noindent \textbf{Takeaway.} \proposedTraining outperforms OFA \wrt ImageNet-1k test accuracy (with 0.5\% better mean pareto accuracy). It improves the accuracy of the smallest subnet by \textbf{1.8\%}. The accuracy improvements come with \textbf{1.8x} training cost reduction compared to OFA. 

\subsection{Ablation Study}
We provide detailed ablation on \proposedTraining components --- \fullModelWarmup, \LRadjust, and \proposedinplaceKD to attribute their benefits.

\begin{figure}[t]
 \begin{subfigure}[t]{0.24\textwidth}
     \includegraphics[width=\columnwidth]{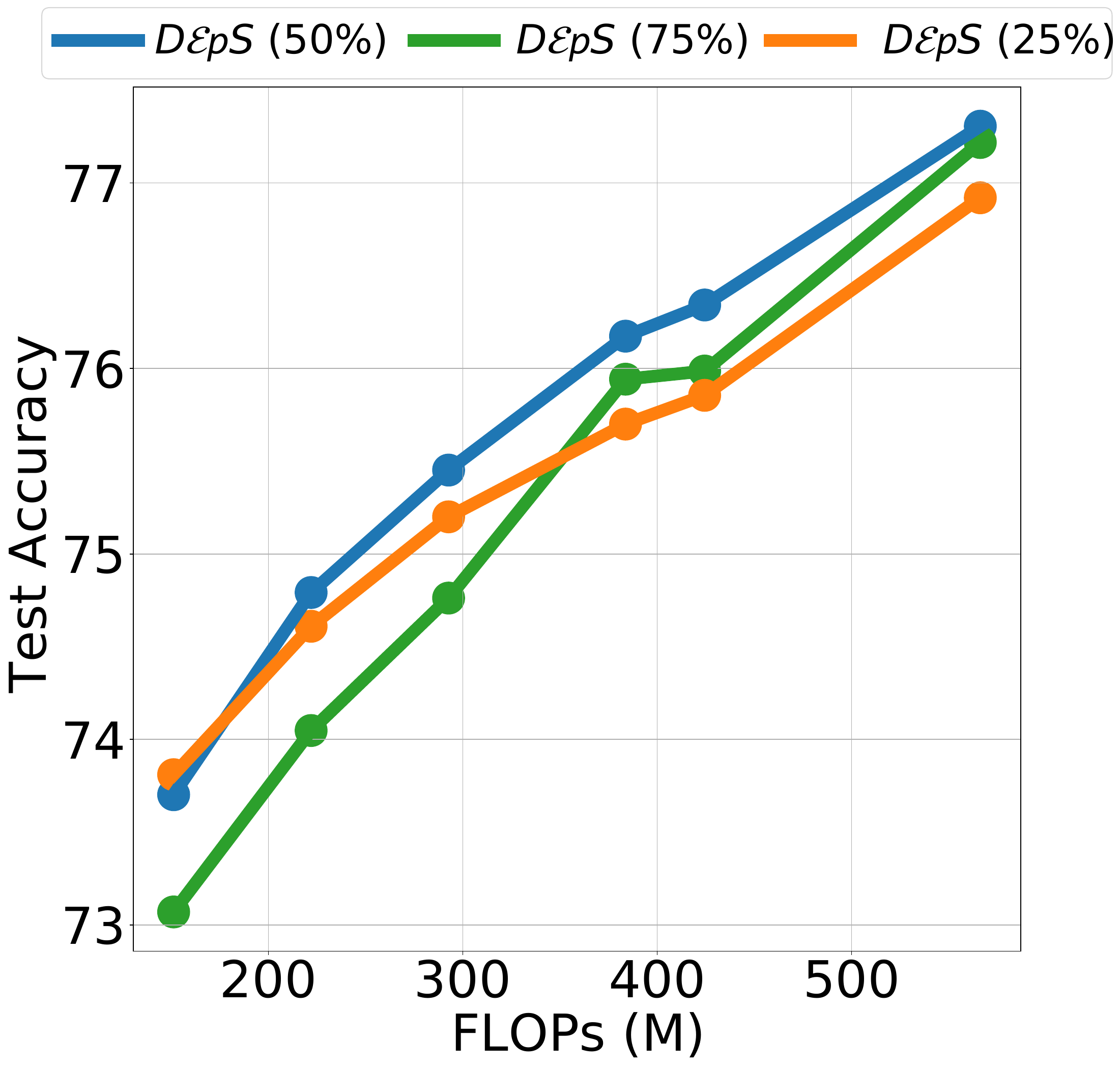}
        \caption{\small \fullModelWarmupSym}
             \label{fig:result:ablations:pfm_warmup}
\end{subfigure}
\begin{subfigure}[t]{0.48\textwidth}
 \includegraphics[width=\columnwidth]{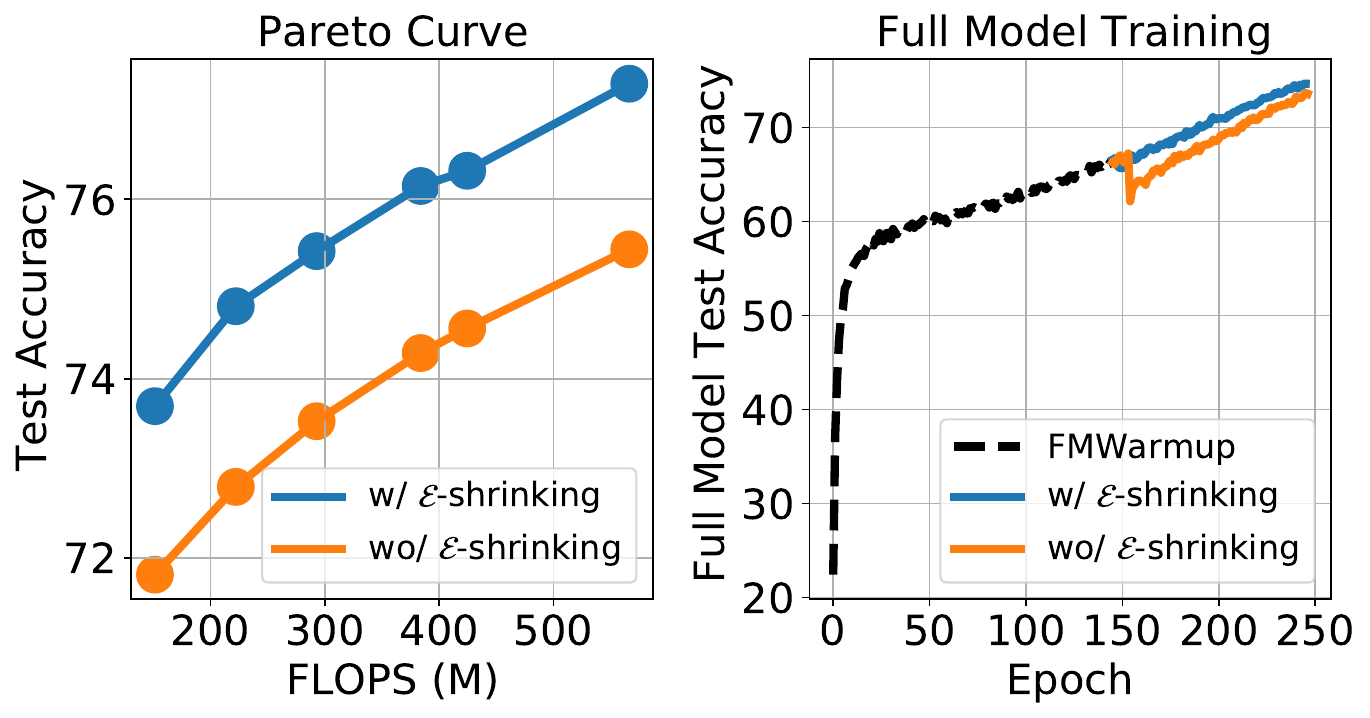}
    \caption{\small \LRadjust}
         \label{fig:result:ablations:LRAdjust}
\end{subfigure}
\begin{subfigure}[t]{0.24\textwidth}
     \includegraphics[width=\columnwidth]{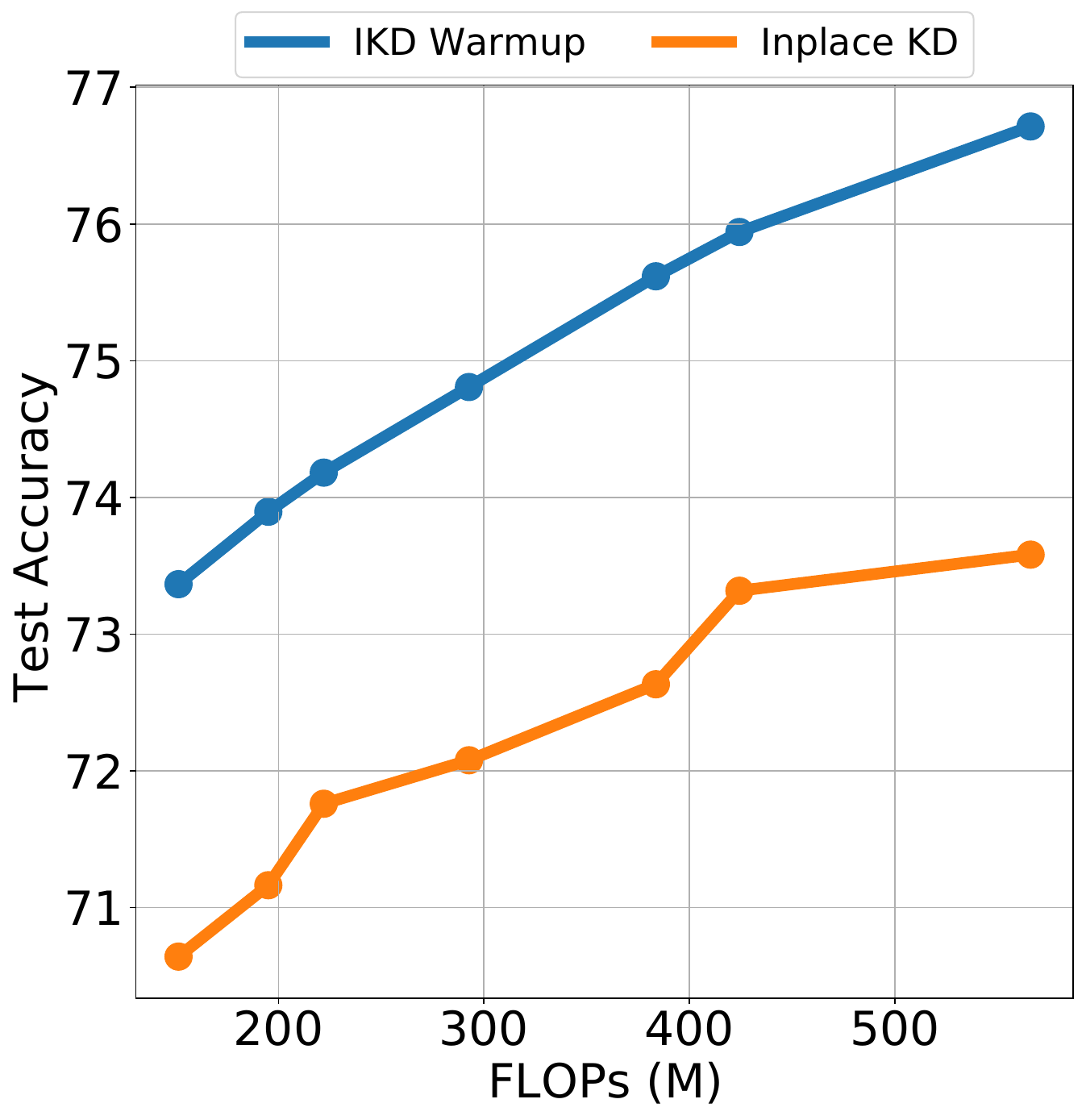}
        \caption{\small Distillation}
             \label{fig:result:ablations:IKDWarmup}
\end{subfigure}

\caption{\small 
    \textbf{ \proposedTraining Ablations.} Three ablations are shown for \proposedTraining --- Full model warmup period (\fullModelWarmupSym), \LRadjust, and Distillation. a) There exists a sweet spot \wrt accuracy (of subnets) in \fullModelWarmupSym (=50\%), b) \LRadjust improves the entire pareto front (left) and prevents drop in accuracy of the full model (right), c) \proposedinplaceKD performs better than Inplace KD as it uses more information from non-target classes (further details are provided in supplementary material).}
    \vspace{-0.2in}
\label{fig:result:ablations}
\end{figure}

\noindent \textbf{Full Model Warmup Period (\fullModelWarmupSym).} In this ablation, we establish the benefits of delayed shrinking as opposed to early or late shrinking. To do this, we configure \proposedTraining to run with different full model warmup periods (\fullModelWarmupSym) -- the time at which shrinking starts in \proposedTraining.  Our goal is to empirically demonstrate the existence of a sweet spot in  \fullModelWarmupSym \wrt accuracy (of subnets). \figref{fig:result:ablations:pfm_warmup} compares the accuracy of best-performing subnets in six different FLOP buckets of three \fullModelWarmupSym periods \{$25$\%,  $50$\%, $75$\%\} on Imagenet-1k dataset. \fullModelWarmupSym=25\%, 75\% represents early and late shrinking respectively. 

\noindent \textit{Takeaway.} \proposedTraining with \fullModelWarmupSym$=50\%$ achieves the best test accuracy across subnets compared to \proposedTraining configured to run with \fullModelWarmupSym$=25\%,75\%$. Hence, a sweet spot exists in \fullModelWarmupSym. 
The existence of a sweet spot demonstrates that both early ($25\%$) or late ($75$\%) shrinking is sub-optimal in training the model family (discussed in \secref{sec:proposed:fm_warmup}). Early shrinking results in sub-optimal accuracy of the larger subnets as training interference occurs very early in the training. While late shrinking causes the specialization of supernet weights to the full model architecture that results in sub-optimal accuracy of smaller subnets ($\approx 1\%$ accuracy degradation around 200 MFLOPs for \fullModelWarmupSym=$75$\% compared to \fullModelWarmupSym=$50$\%).

\noindent \textbf{\epsshrinking.}
We investigate whether an accuracy drop occurs in the full model's accuracy when shrinking is introduced in \proposedTraining and if \epsshrinking prevents it. In this ablation, we run \proposedTraining with and without \epsshrinking and introduce shrinking at 150$^{th}$ epoch while keeping all other training-hyperparameters constant. \figref{fig:result:ablations:LRAdjust} (right) compares \proposedTraining  with and without \epsshrinking on ImageNet-1k top1 test accuracy of the full model over training epochs. \figref{fig:result:ablations:LRAdjust} (left) compares subnets for six different FLOP buckets with and without \LRadjust.

\noindent \textit{Takeaway.} \proposedTraining without \epsshrinking observes a \textbf{2$\%$} drop in full model's accuracy at 150$^{th}$ epoch when the shrinking starts. And, \proposedTraining with \epsshrinking prevents this huge accuracy drop at the start of shrinking that leads to better full model accuracy overall. The prevention of the drop in full model's accuracy demonstrates that \epsshrinking leads to smooth optimization of the full model. \epsshrinking achieves this by incrementally warming up subnets' learning rate at the start of shrinking to avoid sudden changes in the supernet weight (\figref{fig:result:ablations:LRAdjust}, right). \epsshrinking also achieves superior accuracy across the entire FLOP range when compared to the supernet trained without \epsshrinking (\figref{fig:result:ablations:LRAdjust}, left).

\noindent \textbf{\proposedinplaceKD.} 
We assess the benefits of \proposedinplaceKD in this ablation. \proposedinplaceKD performs inplace knowledge distillation from a partially trained full model instead of performing it from the beginning with a randomly initialized full model (inplace KD) as proposed in \cite{slimmable}. Hence, to show benefits of \proposedinplaceKD, we run \proposedTraining with inplace KD and our proposed \proposedinplaceKD . \figref{fig:result:ablations:IKDWarmup} compares \proposedTraining run with \proposedinplaceKD (blue) and inplace KD (orange) on the ImageNet-1k top1 test accuracy of best-performing subnets in seven different FLOP buckets. 

\noindent \textit{Takeaway.} \proposedinplaceKD outperforms inplace KD across all the subnets that cover the supernet's FLOP range on the ImageNet-1k dataset. It is \textbf{3.5\%} and \textbf{2\%}  more accurate at 560 MFLOPs and 150 MFLOPs respectively. This shows that \proposedinplaceKD distills knowledge effectively in once-for-all training as multiple progressively better partially trained  full model transfer their knowledge to smaller subnets (\secref{sec:ikd_warmup}). Inplace KD is not able to provide meaningful knowledge transfer as the full model is under-trained initially.

\section{Conclusion}
\label{sec:conclusion}
\proposedTraining is a training technique that increases the scalability of once-for-all training. \proposedTraining consists of three key components --- \fullModelWarmup that decreases training costs, \LRadjust that keeps the accuracy of the full model on-par with existing works, and \proposedinplaceKD that performs effective knowledge distillation in once-for-all training. \fullModelWarmup's key idea is  to delay the process of shrinking till the full model gets partially trained ($\sim$50\%) to reduce training cost. \LRadjust circumvents accuracy drop in the full model by avoiding  rapid changes in the supernet weights and enabling smooth optimization by incrementally warming up subnets' learning rates. \proposedinplaceKD provides rich knowledge transfer to subnets from multiple partially trained full models that are progressively better \wrt accuracy. \proposedTraining generalizes to different datasets and DNN architecture spaces. It improves the accuracy of smaller subnets, achieves on-par Pareto-optimality, and reduces training cost by upto $2.5$x when compared with existing once-for-all weight-shared training techniques.


%
%
\bibliographystyle{splncs04}
\bibliography{bibs/related_work, bibs/misc, bibs/datasets}

\begin{thebibliography}{10}
\providecommand{\url}[1]{\texttt{#1}}
\providecommand{\urlprefix}{URL }
\providecommand{\doi}[1]{https://doi.org/#1}

\bibitem{tcn}
Bai, S., Kolter, J.Z., Koltun, V.: An empirical evaluation of generic convolutional and recurrent networks for sequence modeling. CoRR  \textbf{abs/1803.01271} (2018), \url{http://arxiv.org/abs/1803.01271}

\bibitem{smart_camera}
Bonnard, J., Abdelouahab, K., Pelcat, M., Berry, F.: On building a cnn-based multi-view smart camera for real-time object detection. Microprocessors and Microsystems  \textbf{77},  103177 (2020)

\bibitem{ofa}
Cai, H., Gan, C., Wang, T., Zhang, Z., Han, S.: Once-for-all: Train one network and specialize it for efficient deployment. In: International Conference on Learning Representations (2020), \url{https://openreview.net/forum?id=HylxE1HKwS}

\bibitem{proxylessnas}
Cai, H., Zhu, L., Han, S.: Proxylessnas: Direct neural architecture search on target task and hardware. arXiv preprint arXiv:1812.00332  (2018)

\bibitem{poorTeacherKD}
Cho, J.H., Hariharan, B.: On the efficacy of knowledge distillation. In: Proceedings of the IEEE/CVF international conference on computer vision. pp. 4794--4802 (2019)

\bibitem{smart_survelliance}
Cob-Parro, A.C., Losada-Guti{\'e}rrez, C., Marr{\'o}n-Romera, M., Gardel-Vicente, A., Bravo-Mu{\~n}oz, I.: Smart video surveillance system based on edge computing. Sensors  \textbf{21}(9), ~2958 (2021)

\bibitem{imagenet1k}
Deng, J., Dong, W., Socher, R., Li, L.J., Li, K., Fei-Fei, L.: Imagenet: A large-scale hierarchical image database. In: 2009 IEEE conference on computer vision and pattern recognition. pp. 248--255. Ieee (2009)

\bibitem{pylot}
Gog, I., Kalra, S., Schafhalter, P., Wright, M.A., Gonzalez, J.E., Stoica, I.: Pylot: A modular platform for exploring latency-accuracy tradeoffs in autonomous vehicles. In: 2021 IEEE International Conference on Robotics and Automation (ICRA). pp. 8806--8813. IEEE (2021)

\bibitem{lr_warmup_2}
Goyal, P., Doll{\'a}r, P., Girshick, R., Noordhuis, P., Wesolowski, L., Kyrola, A., Tulloch, A., Jia, Y., He, K.: Accurate, large minibatch sgd: Training imagenet in 1 hour. arXiv preprint arXiv:1706.02677  (2017)

\bibitem{tinynet}
Han, K., Wang, Y., Zhang, Q., Zhang, W., Xu, C., Zhang, T.: Model rubik’s cube: Twisting resolution, depth and width for tinynets. Advances in Neural Information Processing Systems  \textbf{33},  19353--19364 (2020)

\bibitem{unstructured_pruning_1}
Han, S., Mao, H., Dally, W.J.: Deep compression: Compressing deep neural networks with pruning, trained quantization and huffman coding. arXiv preprint arXiv:1510.00149  (2015)

\bibitem{cnn_search_engine}
Hashemi, H.B., Asiaee, A., Kraft, R.: Query intent detection using convolutional neural networks. In: International conference on web search and data mining, workshop on query understanding (2016)

\bibitem{lr_warmup_1}
He, K., Zhang, X., Ren, S., Sun, J.: Deep residual learning for image recognition. In: Proceedings of the IEEE conference on computer vision and pattern recognition. pp. 770--778 (2016)

\bibitem{vanillakd}
Hinton, G., Vinyals, O., Dean, J.: Distilling the knowledge in a neural network. arXiv preprint arXiv:1503.02531  (2015)

\bibitem{mbv3}
Howard, A., Sandler, M., Chu, G., Chen, L.C., Chen, B., Tan, M., Wang, W., Zhu, Y., Pang, R., Vasudevan, V., et~al.: Searching for mobilenetv3. In: Proceedings of the IEEE/CVF international conference on computer vision. pp. 1314--1324 (2019)

\bibitem{quantization_1}
Hubara, I., Courbariaux, M., Soudry, D., El-Yaniv, R., Bengio, Y.: Binarized neural networks. Advances in neural information processing systems  \textbf{29} (2016)

\bibitem{squeezenet}
Iandola, F.N., Han, S., Moskewicz, M.W., Ashraf, K., Dally, W.J., Keutzer, K.: Squeezenet: Alexnet-level accuracy with 50x fewer parameters and< 0.5 mb model size. arXiv preprint arXiv:1602.07360  (2016)

\bibitem{nvidia_jetson}
Inc., N.: Nvidia jetson. \url{https://www.nvidia.com/en-in/autonomous-machines/embedded-systems/}, [Accessed 13-May-2023]

\bibitem{nvidia_v100}
Inc, N.: Nvidia v100. \url{https://www.nvidia.com/en-in/data-center/v100/}, [Accessed 13-May-2023]

\bibitem{quantization_2}
Jacob, B., Kligys, S., Chen, B., Zhu, M., Tang, M., Howard, A., Adam, H., Kalenichenko, D.: Quantization and training of neural networks for efficient integer-arithmetic-only inference. In: Proceedings of the IEEE conference on computer vision and pattern recognition. pp. 2704--2713 (2018)

\bibitem{cifar}
Krizhevsky, A., Hinton, G., et~al.: Learning multiple layers of features from tiny images  (2009)

\bibitem{structured_pruning_1}
Li, H., Kadav, A., Durdanovic, I., Samet, H., Graf, H.P.: Pruning filters for efficient convnets. arXiv preprint arXiv:1608.08710  (2016)

\bibitem{structured_pruning_2}
Lin, M., Ji, R., Wang, Y., Zhang, Y., Zhang, B., Tian, Y., Shao, L.: Hrank: Filter pruning using high-rank feature map. In: Proceedings of the IEEE/CVF conference on computer vision and pattern recognition. pp. 1529--1538 (2020)

\bibitem{sgdr}
Loshchilov, I., Hutter, F.: Sgdr: Stochastic gradient descent with warm restarts. arXiv preprint arXiv:1608.03983  (2016)

\bibitem{structured_pruning_3}
Luo, J.H., Wu, J., Lin, W.: Thinet: A filter level pruning method for deep neural network compression. In: Proceedings of the IEEE international conference on computer vision. pp. 5058--5066 (2017)

\bibitem{self_driving}
Ouyang, Z., Niu, J., Liu, Y., Guizani, M.: Deep cnn-based real-time traffic light detector for self-driving vehicles. IEEE transactions on Mobile Computing  \textbf{19}(2),  300--313 (2019)

\bibitem{real2019regularized}
Real, E., Aggarwal, A., Huang, Y., Le, Q.V.: Regularized evolution for image classifier architecture search. In: Proceedings of the aaai conference on artificial intelligence. vol.~33, pp. 4780--4789 (2019)

\bibitem{imagenet}
Russakovsky, O., Deng, J., Su, H., Krause, J., Satheesh, S., Ma, S., Huang, Z., Karpathy, A., Khosla, A., Bernstein, M., Berg, A.C., Fei-Fei, L.: {ImageNet Large Scale Visual Recognition Challenge}. International Journal of Computer Vision (IJCV)  \textbf{115}(3),  211--252 (2015). \doi{10.1007/s11263-015-0816-y}

\bibitem{compofa}
Sahni, M., Varshini, S., Khare, A., Tumanov, A.: Comp{\{}ofa{\}} {\textendash} compound once-for-all networks for faster multi-platform deployment. In: International Conference on Learning Representations (2021), \url{https://openreview.net/forum?id=IgIk8RRT-Z}

\bibitem{unstructured_pruning_2}
Sanh, V., Wolf, T., Rush, A.: Movement pruning: Adaptive sparsity by fine-tuning. Advances in Neural Information Processing Systems  \textbf{33},  20378--20389 (2020)

\bibitem{unstructured_pruning_3}
Sun, W., Zhou, A., Stuijk, S., Wijnhoven, R., Nelson, A.O., Corporaal, H., et~al.: Dominosearch: Find layer-wise fine-grained n: M sparse schemes from dense neural networks. Advances in neural information processing systems  \textbf{34},  20721--20732 (2021)

\bibitem{mnasnet}
Tan, M., Chen, B., Pang, R., Vasudevan, V., Sandler, M., Howard, A., Le, Q.V.: Mnasnet: Platform-aware neural architecture search for mobile. In: Proceedings of the IEEE/CVF conference on computer vision and pattern recognition. pp. 2820--2828 (2019)

\bibitem{efficientnet}
Tan, M., Le, Q.: Efficientnet: Rethinking model scaling for convolutional neural networks. In: International conference on machine learning. pp. 6105--6114. PMLR (2019)

\bibitem{imagenet100}
Tian, Y., Krishnan, D., Isola, P.: Contrastive multiview coding. In: Computer Vision--ECCV 2020: 16th European Conference, Glasgow, UK, August 23--28, 2020, Proceedings, Part XI 16. pp. 776--794. Springer (2020)

\bibitem{fbnet_v2}
Wan, A., Dai, X., Zhang, P., He, Z., Tian, Y., Xie, S., Wu, B., Yu, M., Xu, T., Chen, K., et~al.: Fbnetv2: Differentiable neural architecture search for spatial and channel dimensions. In: Proceedings of the IEEE/CVF conference on computer vision and pattern recognition. pp. 12965--12974 (2020)

\bibitem{partialKD}
Wang, C., Yang, Q., Huang, R., Song, S., Huang, G.: Efficient knowledge distillation from model checkpoints. In: Oh, A.H., Agarwal, A., Belgrave, D., Cho, K. (eds.) Advances in Neural Information Processing Systems (2022), \url{https://openreview.net/forum?id=0ltDq6SjrfW}

\bibitem{quantization_3}
Wang, L., Dong, X., Wang, Y., Liu, L., An, W., Guo, Y.: Learnable lookup table for neural network quantization. In: Proceedings of the IEEE/CVF conference on computer vision and pattern recognition. pp. 12423--12433 (2022)

\bibitem{fbnet}
Wu, B., Dai, X., Zhang, P., Wang, Y., Sun, F., Wu, Y., Tian, Y., Vajda, P., Jia, Y., Keutzer, K.: Fbnet: Hardware-aware efficient convnet design via differentiable neural architecture search. In: Proceedings of the IEEE/CVF Conference on Computer Vision and Pattern Recognition. pp. 10734--10742 (2019)

\bibitem{inplacekd}
Yu, J., Huang, T.S.: Universally slimmable networks and improved training techniques. In: Proceedings of the IEEE/CVF international conference on computer vision. pp. 1803--1811 (2019)

\bibitem{bignas}
Yu, J., Jin, P., Liu, H., Bender, G., Kindermans, P.J., Tan, M., Huang, T., Song, X., Pang, R., Le, Q.: Bignas: Scaling up neural architecture search with big single-stage models. In: Vedaldi, A., Bischof, H., Brox, T., Frahm, J.M. (eds.) Computer Vision -- ECCV 2020. pp. 702--717. Springer International Publishing, Cham (2020)

\bibitem{slimmable}
Yu, J., Yang, L., Xu, N., Yang, J., Huang, T.: Slimmable neural networks. arXiv preprint arXiv:1812.08928  (2018)

\bibitem{zoph2018learning}
Zoph, B., Vasudevan, V., Shlens, J., Le, Q.V.: Learning transferable architectures for scalable image recognition. In: Proceedings of the IEEE conference on computer vision and pattern recognition. pp. 8697--8710 (2018)

\end{thebibliography}

\end{document}